%% file: main.tex
\documentclass[10pt,twocolumn,letterpaper]{article}

\usepackage{cvpr}              %

\input{preamble}

\definecolor{cvprblue}{rgb}{0.21,0.49,0.74}
\usepackage[pagebackref,breaklinks,colorlinks,citecolor=cvprblue]{hyperref}

\def\paperID{6} %
\def\confName{CVPRW}
\def\confYear{2024}

\title{Blind Localization and Clustering of Anomalies in Textures}

\author{Andrei-Timotei Ardelean\qquad Tim Weyrich\\[0.75ex]
Friedrich-Alexander-Universität Erlangen-Nürnberg\\[0.25ex]
{\tt\small \{timotei.ardelean, tim.weyrich\}@fau.de}
}

\newcommand{\myparagraph}[2][\hspace{0.5em}]{\smallskip\noindent\mbox{\textbf{#2}}#1}

\newcommand{\kmeans}{$k$-means\xspace}
\newcommand{\semisup}{normality-supervised\xspace}
\newcommand{\Semisup}{Normality-supervised\xspace}

\begin{document}
\maketitle

\input{sec/0_abstract}    
\begin{figure}[b]
  \centering%
  \vspace*{-6.5ex}%
  \includegraphics[width=0.94\linewidth]{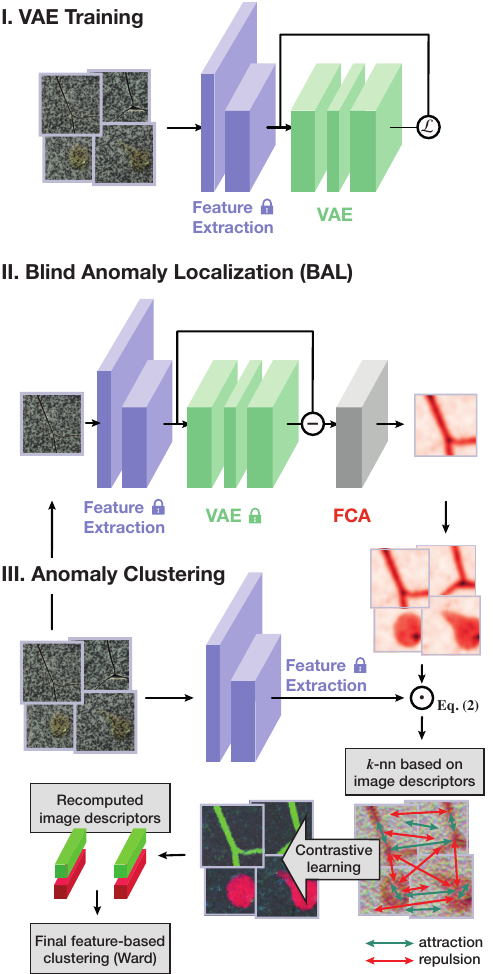}\\[-0.5ex]
  \caption{\label{fig:pipeline}%
    System overview. We train a VAE on features extracted from
    the input set, and apply FCA~\cite{ardelean2024high} on the residuals to obtain anomaly
    maps in a blind setting. These maps are used to mine
    positive and negative pairs for contrastive learning. The resulting improved image-level descriptors are then hierarchically
    clustered.\vspace*{-0.75ex}}
\end{figure}
\input{sec/1_introduction}
\input{sec/2_related}
\input{sec/3_method}
\input{sec/4_experiments}

\input{sec/5_conclusion}

{
    \small
    \bibliographystyle{ieeenat_fullname}
    \bibliography{main}
}

\input{sec/X_suppl}

\end{document}

%% file: preamble.tex
\usepackage[dvipsnames]{xcolor}

\usepackage{microtype}
\usepackage{newtxtext}
\usepackage{newtxmath}
\usepackage{multirow}
\usepackage[accsupp]{axessibility}

%% file: sec/0_abstract.tex
\begin{abstract}
  Anomaly detection and localization in images is a growing field in computer vision. In this area, a seemingly understudied problem is anomaly clustering, i.e.,
  identifying and grouping different types of anomalies
  in a fully unsupervised manner. In this work, we propose a novel method for clustering anomalies in largely stationary images (textures) in a blind setting. That is, the input consists of normal and anomalous images without distinction and without labels. What contributes to the difficulty of the task is that anomalous regions are often small and may present only subtle changes in appearance, which can be easily overshadowed by the genuine variance in the texture. Moreover, each anomaly type may have a complex appearance distribution. 
  We introduce a novel scheme for solving this task using a combination of blind anomaly localization and contrastive learning. By identifying the anomalous regions with high fidelity, we can restrict our focus to
  those
  regions of interest; then, contrastive learning is employed to increase the separability of different anomaly types and reduce the intra-class variation. Our experiments show that the proposed solution yields significantly better results compared to prior work, setting a new state of the art.
  Project page: {\footnotesize\texttt{\href{https://reality.tf.fau.de/pub/ardelean2024blind.html}{reality.tf.fau.de/pub/ardelean2024blind.html}}}.
\end{abstract}

%% file: sec/1_introduction.tex
\section{Introduction}
\label{sec:intro}

The detection of anomalies is an important problem, found in applications from widely different domains such as medicine, manufacturing, security, finance, etc.
The problem is of great interest to the computer vision community as the anomalies are often observed in images or videos.
In this work, we focus on identifying structural anomalies, generally found in textured surfaces, rather than generic object anomalies.
Depending on the application, there are several tasks covered by the overarching title of anomaly detection, such as image-level detection, pixel-level detection (localization), and clustering of anomalies.
These tasks can further be divided depending on the level of supervision that is available.

Traditionally, there are three main categories: supervised, semi-supervised, and unsupervised methods~\cite{chalapathy2019deep, chandola2009anomaly}.
Supervised methods assume the existence of a training dataset with labeled anomalies,
a setting resembling supervised classification
with unbalanced classes.
Semi-supervised detection of anomalies only requires a curated set of normal samples (no anomalies) for training.
Unsupervised methods do not require any labels and do not use a separate training set, but directly detect anomalies based on the data distribution. This setting has also been named \textbf{blind anomaly detection}~\cite{zhang2023s} and fully unsupervised learning~\cite{cordier2022sroc}.
Notably, a large number of relevant works~\cite{roth_towards_2022, defard2021padim, lee2022cfa, gudovskiy2022cflow}, including surveys~\cite{liu2024deep, tao2022deep}, use the term of unsupervised anomaly detection while referring to the second category described above.
To avoid the ambiguity regarding these terms, we refer to the training-set-free category as \emph{blind}, following~\cite{zhang2023s}, and use the term \emph{\semisup} for methods that use a curated set of anomaly-free images.

Anomaly clustering aims to not only detect outliers but also group the anomalies found among a set of elements into semantically coherent clusters.
Such a system could greatly improve the interpretability of anomaly detection systems and allow more control over the results; for example, when only certain anomaly types are relevant for the end-application, the others can be filtered out.
We address the problem of blind anomaly clustering in images, specifically textures, which are (mostly) stationary.
We propose to solve anomaly clustering in two stages. 
Firstly, the anomalous regions are identified 
regardless of the type, essentially performing blind anomaly localization (BAL).
Secondly, using contrastive learning, we map the original images to a new feature space where different anomaly types can more easily be separated.

In the BAL setting, the input consists of a set of images which may or may not contain anomalies, without any labels. 
Applying an existing \semisup method~\cite{roth_towards_2022, li2021cutpaste} would be suboptimal due to the (possibly high level of) anomaly-contamination~\cite{yoon2022srr, cordier2022sroc}.
At the same time, zero-shot (single-image) methods such as FCA~\cite{ardelean2024high} and Aota \etal~\cite{aota_zero-shot_2023} would not leverage the information from all images.
Therefore, we propose a novel approach specifically for BAL. We harness the benefits of the zero-shot FCA and combine it with a variational autoencoder (VAE) trained on the entire input set, injecting global information in each image.

After localizing the anomalies, one can aggregate the anomalous features and apply a classic clustering algorithm on the pooled descriptors, such as \kmeans, hierarchical clustering, spectral clustering, etc.
However, we find that due to the intrinsic variation of anomalous features, coupled with the imperfect localizations from BAL, the anomaly classes are hard to separate.
We address this by performing contrastive learning, mining positive and negative pairs using the computed anomaly maps.
This greatly improves the clustering performance and increases its stability.

To summarize, our contributions are as follows:
\begin{itemize}
    \item We are the first to make the connection between blind anomaly localization (BAL) and anomaly clustering by formulating the problem of anomaly clustering as a two-stage process: BAL, followed by feature fine-tuning.
    \item We derive a novel solution for BAL, which applies the recent zero-shot method Feature Correspondence Analysis (FCA)~\cite{ardelean2024high} to residual maps obtained using a variational autoencoder.
    \item We propose a way to fine-tune features for anomaly clustering through contrastive learning, leveraging the anomaly maps predicted by BAL.
\end{itemize}

%% file: sec/2_related.tex
\section{Related Work}
\label{sec:related}

Blind anomaly localization and clustering is a complex problem, with strong connections to several directions of research in computer vision.
In this section, we briefly revise the tasks adjacent to our problem formulation and how the existing solutions relate to anomaly clustering.

\myparagraph{\Semisup anomaly detection.}
Most of the prominent progress in anomaly detection in images consists of \semisup methods; \ie, methods which rely on a set of normal images, used as reference to compare against at test time.
There are several classes of methods that follow this pattern. 
Reconstruction-based methods generally use autoencoders~\cite{Bergmann2018ImprovingUD, Youkachen2019DefectSO}, VAEs~\cite{an2015variational, Baur2018DeepAM, zimmerer2019unsupervised, zhou2020unsupervised}, or GANs~\cite{Akay2019SkipGANomalySC, Baur2018DeepAM, Schlegl2017UnsupervisedAD} which learn to reproduce normal images.
During inference, comparing the reconstruction with the input image should yield higher errors in anomalous regions, since the network was not 
trained to reconstruct such elements.
Another set of methods relies on pretrained feature extractors, modeling normality in different ways including feature banks~\cite{roth_towards_2022}, per-pixel gaussian distributions~\cite{defard2021padim}, $k$-nearest neighbors search~\cite{bergman2020deep, cohen2020sub}, etc.
Modeling the features from the normal set allows assigning anomaly scores by measuring how well the model explains a new sample.
Other approaches have also been proven competitive, such as self-supervised learning (synthesizing artificial anomalies)~\cite{li2021cutpaste, zavrtanik2021draem}, flow-based models~\cite{gudovskiy2022cflow}, contrastive learning~\cite{lee2022cfa}, etc.
These methods show impressive results on the usual benchmarks; however, it has been shown that in general the performance degrades significantly when the normal set is contaminated (contains some anomalous images)~\cite{yoon2022srr, cordier2022sroc, zhang2023s}.

\myparagraph{Blind anomaly detection.}
The main response to the problem of anomaly contamination for \semisup methods is represented by training-set refinement. 
SRR~\cite{yoon2022srr}, SROC~\cite{cordier2022sroc}, and STKD~\cite{liu2022unsupervised} present different ways to detect outliers in the anomaly-contaminated training set and eliminate them. 
The refined set can then be used by a \semisup method to build the notion of normality.
The results of these approaches have been shown for small amounts of contamination (up to 20\%), with degrading performance for higher factors. On the other hand, our task concerns a generic blind anomaly setting, where the number of anomalous images can be much higher (up to 75\% in our experiments).
Patel \etal~\cite{patel2023self} developed a solution for such cases, where the ratio of anomalous samples is high. Instead of eliminating the outlying images, only the anomalous regions are masked, allowing the method to leverage the normalcy information available in these images.
Another approach explicitly designed for blind anomaly detection was recently proposed by Zhang \etal~\cite{zhang2023s}. 
The method builds on PatchCore~\cite{roth_towards_2022}, addressing the challenge of creating an anomaly-robust feature bank.
Our approach takes a different path, \ie, combining variational feature reconstruction with a zero-shot detection method.

\myparagraph{Zero-shot anomaly localization}
A distinct category of supervision in anomaly detection problems consists of zero-shot methods.
This scenario assumes that each image is analyzed in isolation; therefore, it heavily relies on priors and/or presumed stationarity of the input image.
Strong priors are leveraged by WinCLIP~\cite{Jeong_2023_CVPR}, SAA~\cite{cao2023segment}, and April-GAN~\cite{chen2023zero} in the form of pretrained visual-language models, using text prompts to identify anomalous regions.
On the other hand, the methods of Aota \etal~\cite{aota_zero-shot_2023} and Ardelean and Weyrich (FCA)~\cite{ardelean2024high} target textured images. Here, the anomalies are marked as the regions that break the overall homogeneity of the image. More concretely, FCA computes the anomaly score of a pixel by computing its contribution to the Wasserstein distance between the local (patch) and global (image) distributions.
These methods can theoretically be directly applied to blind anomaly detection; however, they would not leverage all available information, suggesting they are suboptimal for this task.

\myparagraph{Deep Image Clustering}
Consistently with prior work~\cite{sohn2023anomaly, lee2023selformaly}, we assume that each image has only one type of anomaly.
This makes the task similar to generic image clustering, tackled by various methods such as IIC~\cite{ji2019iic}, GATCluster~\cite{niu2020gatcluster}, SCAN~\cite{van2020scan}, and SPICE~\cite{niu2022spice}.
Nonetheless, anomaly clustering has two peculiarities which make it more challenging.
Firstly, the classes are naturally unbalanced, and we assume no prior knowledge on how rare a specific anomaly type is compared to other types or the normal class.
Secondly, the anomalies can be small and subtle, making it difficult to cluster the images when approached holistically~\cite{sohn2023anomaly}.

\myparagraph{Unsupervised semantic segmentation.}
The combination of clustering and localization of anomalies can also be seen as an instance of unsupervised semantic segmentation.
For example, methods such as PiCIE~\cite{cho2021picie}, STEGO~\cite{hamilton2022unsupervised} and HP~\cite{seong2023leveraging} identify semantically similar regions in different images.
Our anomaly-targeted, margin-based contrastive learning algorithm bears similarities with the correspondence distillation formulation of Hamilton \etal~\cite{hamilton2022unsupervised}.
However, the method we propose is adapted to the specifics of anomaly clustering, which gives it a performance edge.

%% file: sec/3_method.tex
\section{Method}
\label{sec:method}

We propose a system for clustering anomalies as a two-stage process. 
The first component performs blind anomaly localization (BAL) by extending the zero-shot feature correspondence analysis (FCA) algorithm of Ardelean and Weyrich~\cite{ardelean2024high} to leverage the properties of reconstruction-based anomaly detection methods.
The second component refines the texture descriptors used for clustering to make the anomalous classes more separable.
Finally, the instances are clustered using classic feature clustering based on image-level embeddings.
An overview of the pipeline is presented in \Cref{fig:pipeline}.

\subsection{Blind Anomaly Localization}

The task of blind anomaly localization can be formulated as a function that takes as input a set of images $\{I_i\}_{i=1}^N, I_i \in \mathbb{R}^{H\times W \times 3}$ and returns an anomaly map ($A_i \in \mathbb{R}^{H\times W}$) for each image. 
Note that the set of images has mixed normal and anomalous images, and no labels are available.

Applying a zero-shot method for each image in isolation may work well for simple, stationary textures, but fails when the normal class is more complex,
so that
information from multiple images is required to determine anomalies. 
In order to homogenize the distribution in the normal regions using the available data, we employ a Variational Autoencoder (VAE) trained to reconstruct the input. 
Instead of operating on images directly, we use a feature extractor $F$ consisting of the first layers of a neural network pretrained on ImageNet~\cite{deng2009imagenet}; the variational reconstruction is performed on the extracted features.
We observe that using $1\times 1$ convolutions is sufficient, as the neural features already capture the local information of the patch in the RGB image; this also allows the network to be trained very fast.
At inference, we encode the input features (\ie, $\mu_i, \sigma_i = \textit{VAE}_E(F(I_i))$) and use the predicted embedding mean ($\mu_i$) for decoding. 
The residual between the input and the reconstruction is then used as input to FCA using the default parameters ($\sigma_p$ = 3.0 and $\sigma_s = 1.0$), yielding the final \Cref{eq:bal}:
\begin{equation}
    A_i = \textit{FCA}(F(I_i) - \textit{VAE}_D(\mu_i)) \;.
\label{eq:bal}
\end{equation}
The union of the two elements is elegant and straightforward, and it works surprisingly well in practice.
In the following, we give another intuition for the connection, and we empirically demonstrate this synergy in the experiments section.

A prominent class of methods for \semisup anomaly detection is represented by the reconstruction-based approaches, \eg, a Variational Autoencoder trained to reconstruct normal images. 
At inference time, given an anomalous image, the network will mostly succeed in recovering normal regions. 
On the other hand, anomalies will not be correctly represented by the network since they were not observed during training.
Therefore, the mismatch between the original image and the reconstructions can be used as an anomaly score. 
One can simply use the $L^1$ or $L^2$ norm of the difference~\cite{Baur2018DeepAM, chen2018deep}, or more involved measures such as SSIM~\cite{Bergmann2018ImprovingUD}, or Rec-grad~\cite{zimmerer2019unsupervised, zhou2020unsupervised} for obtaining the final anomaly map. 
As observed in~\cite{patel2023self, Baur2018DeepAM}, VAE-based methods have a certain natural robustness to anomaly contamination and generally perform better than simpler autoencoders. 
This is because the \emph{normal} pixels constitute the overwhelming majority, and the regularization of the latent space forces the network to focus on minimizing the error on this majority. 
While the magnitudes of the residuals between the input and reconstruction can be relatively noisy~\cite{Bergmann2018ImprovingUD}, the structure of the errors provides an additional discriminatory signal for localizing the anomalies.
Concretely, we observe that identifying abnormal structures in the feature residuals is equivalent to a zero-shot anomaly localization task where the input consists of the difference between the original features and the reconstruction.

\subsection{Clustering}
From the previous step, we now have for each image an anomaly map $A_i \in \mathbb{R}^{H\times W}$ and the extracted feature map $F_i = F(I_i), F_i \in \mathbb{R}^{H\times W\times C}$. 
When performing clustering, we use the same assumption as Sohn \etal \cite{sohn2023anomaly} and consider each image to subsume at most one type of anomaly. 
In this case, the clustering of anomalies can be made by clustering the respective images in the input set (with one extra cluster for the images that do not present any anomalies). 
To this end, Sohn \etal aggregate the features using the anomaly maps to obtain one descriptor per image and then use agglomerative clustering on the summative descriptors.

Due to the high variability in the appearance of anomalies of the same type, it is hard to separate clusters based on aggregate descriptors in the original space (\ie, the feature space of the pretrained network). 
Therefore, we propose to fine-tune descriptors by adding an additional $1\times 1$ convolutional head after the layers extracted from the pretrained network $F$, which are kept frozen. 
The added convolutions are trained using a contrastive learning regime as depicted in part III of \Cref{fig:pipeline}.

To obtain positive and negative pairs for contrastive learning, we first binarize the anomaly maps using a threshold $t$ computed based on the distribution of the anomaly scores. 
In practice, this threshold can be selected using validation images or based on additional information regarding the amount of anomalies expected in the dataset. 
In our experiments, we used a generic heuristic described in more detail in the supplementary material, ~\Cref{sec:threshold}.

For an image $i$, let $S_i \in \mathbb{R}^{K_i\times C}$ denote the set of anomalous features (where the anomaly value is larger than the threshold), \ie, $S_i = \{H(F_i)^{xy}) \mid A_i^{xy} > t \}$, and $\bar{S}_i \in \mathbb{R}^{(H\cdot W-K_i)\times C}$ the set of normal features. 
$H$ represents the added trainable layers. 
One can now form positive pairs from $S_i\times S_i$ and $\bar{S}_i\times \bar{S}_i$, as well as negative pairs from $S_i\times \bar{S}_i$.
However, this setup only ensures consistency between normal and anomalous features at the level of each image and will not preserve the relationship between different types of anomalies.
To find positive and negative pairs between different images we first find the $k$-nearest neighbors of each image based on a cumulative feature, weighted by the anomaly map, calculated as\vspace*{-0.333\baselineskip}
\begin{equation}
    D_i = \sum_{xy}\Big(F_i^{xy}\frac{e^{A_i^{xy}\tau^{-1}}}{\sum_{xy}e^{A_i^{xy}\tau^{-1}}}\Big) \;.
\label{eq:descriptor}
\end{equation}
Using the image descriptors $\{D_i\}_{i=1}^N$, we find the $k$-nearest neighbors $\mathcal{N}(i) \in \mathbb{N}^k$, and we sample $k$ non-neighbors from the bottom $0.5$-quantile of the distances: $\mathcal{C}(i) \in \mathbb{N}^k$.
From these image-level relations, we pool the anomalous and normal features to obtain collective sets of neighboring anomalies ($P_i = S_i \cup \bigcup_{j \in \mathcal{N}(i)}S_j$), neighboring normals ($\bar{P}_i = \bar{S}_i \cup \bigcup_{j \in \mathcal{N}(i)}\bar{S}_j$), and non-neighboring anomalies ($C_i = \bigcup_{j \in \mathcal{C}(i)}S_j$).
We can now form more suitable positive pairs from $S_i\times P_i$ and $\bar{S}_i\times \bar{P}_i$ and negative pairs from $S_i\times \bar{P}_i$ and $P_i\times C_i$.
Finally, the described pairs are processed using the contrastive loss formulation of Hadsell \etal~\cite{hadsell2006dimensionality} to optimize $H$.

After training $H$, we obtain clusters by computing (again) image-level descriptors as in Equation~\ref{eq:descriptor}, but using $H(F_i)$ instead of $F_i$ directly.
The descriptors are then processed by an off-the-shelf feature-clustering method; we use agglomerative clustering with Ward linkage in our experiments.

\subsection{Implementation Details}
The variational autoencoder uses 4 convolutional $1\times 1$ layers to compute $\mu_i$ and $\sigma_i$ in the encoder and contains 3 layers in the decoder.
All convolutions are followed by ReLu activations.
The number of feature channels in each intermediate layer matches the input (\ie, $512$) except for the latent bottleneck where dimension is $128$. 
For each instance, the input features are rescaled between $0$ and $1$ as in~\cite{ardelean2024high}.
The VAE is trained for 10k iterations, using AdamW as an optimizer with learning rate of $10^{-4}$ and $0.1$ weight decay; the optimization takes about one minute.

For simplicity, all images are resized to $512\times 512$ before feature extraction. Following \cite{ardelean2024high, aota_zero-shot_2023}, we use the output of the second convolutional block of a pretrained WideResnet-50~\cite{BMVC2016_87}. 
The anomaly maps predicted by FCA are smoothed using a Gaussian kernel with $\sigma=1.0$, and the features are smoothed with $\sigma=2.0$ and centered to have the mean equal $0$ before computing \Cref{eq:descriptor}. 
Similar in effect with the average pooling in Sohn \etal~\cite{sohn2023anomaly}, the blurring reduces high-frequency noise and facilitates clustering.
Following \cite{ardelean2024high}, we discard the borders of the images as the anomaly scores are not reliable near the edges.

\begin{table*}[ht]
  \centering%
  \newcommand{\cs}{\quad}%
  \begin{tabular}{l@{\cs\cs}ccc@{\cs\cs}ccc@{\cs\cs}ccc@{\cs}}
    \toprule
    \multicolumn{1}{l}{Method} &
    \multicolumn{3}{c@{\cs\cs\cs}}{MVTec~AD textures} &
    \multicolumn{3}{c@{\cs\cs}}{MTD} &
    \multicolumn{3}{c@{\cs\cs}}{Leaves}\\[0.5ex]
    
    &
    NMI & ARI & $F_1$ &
    NMI & ARI & $F_1$ &
    NMI & ARI & $F_1$ \\

    \midrule
    
    \color{Gray} SelFormaly~\cite{lee2023selformaly} & \color{Gray} 0.743 & \color{Gray} 0.675 & \color{Gray} 0.795
    & \color{Gray}-- & \color{Gray}-- & \color{Gray}--
    & \color{Gray}-- & \color{Gray}-- & \color{Gray}-- \\
    
    \cmidrule(r{2.25em}){2-4}%
    \cmidrule(r{2em}){5-7}%
    \cmidrule(r{1em}){8-10}

    SPICE~\cite{niu2022spice}
    & 0.268 & 0.079 & 0.174
    & 0.028 & 0.014 & 0.309
    & 0.351 & 0.272 & 0.429 \\

    SCAN~\cite{van2020scan}
    & 0.277 & 0.153 & 0.335
    & 0.071 & 0.029 & 0.282
    & 0.394 & 0.356 & 0.526 \\

    STEGO~\cite{hamilton2022unsupervised}
    & 0.389 & 0.271 & 0.459
    & 0.056 & 0.111 & 0.586
    & 0.055 & 0.021 & 0.344 \\

    Average~\cite{sohn2023anomaly}
    & 0.273 & 0.123 & 0.402
    & 0.065 & 0.024 & 0.289
    & 0.341 & 0.359 & 0.519 \\

    Max Hausdorff~\cite{sohn2023anomaly}
    & 0.625 & 0.534 & 0.708
    & 0.193 & 0.112 & 0.381
    & 0.546 & 0.605 & 0.780 \\

    Weighted Avg~\cite{sohn2023anomaly}
    & 0.674 & 0.601 & 0.707
    & 0.179 & 0.120 & 0.346
    & 0.459 & 0.462 & 0.630 \\

    Ours
    & \textbf{0.790} & \textbf{0.716} & \textbf{0.806}
    & \textbf{0.221} & \textbf{0.392} & \textbf{0.670}
    & \textbf{0.712} & \textbf{0.736} & \textbf{0.829} \\

    \bottomrule
  \end{tabular}\\[-0.25\baselineskip]
  \caption{\label{tab:main_results}
    Comparison between different methods applied to anomaly clustering.
    For all metrics a higher value is better.
  }
\end{table*}

The network $H$ trained using contrastive learning consists of two $1\times 1$ convolutions with $512$ channels, followed by $L^2$ normalization. The layers are trained using AdamW($\textit{lr}=5\cdot 10^{-4}, \textit{weight decay}=0.01$) for $10$ epochs for each texture class.
We take $k=3$ neighbors and use a margin of $0.5$.

The algorithm is not particularly sensitive to the aforementioned arguments. The most important parameter of our approach is the temperature $\tau$ (see \Cref{eq:descriptor}) which is set to $0.002$ in our experiments unless noted otherwise. We study the behaviour of this parameter in~\Cref{sub:sensitivity}.

%% file: sec/4_experiments.tex
\section{Experiments}
\label{sec:experiments}

To validate the utility of our contributions we compare the proposed anomaly clustering system to the current %
state of the art%
. We additionally assess the importance of each component through an extensive ablation study.

\subsection{Datasets}
\label{sub:datasets}
Since we are targeting anomaly clustering in textures, we use some of the few datasets that contain largely homogeneous materials and present various (labeled) types of anomalies. 
Note that the anomaly labels are needed for evaluation but never seen during processing.
The datasets used are MVTec AD~\cite{bergmann2019mvtec, bergmann2021mvtec}, magnetic tile defect (MTD)~\cite{huang2020surface}, and coffee leaves dataset (Leaves)~\cite{ESGARIO2020105162}.
As mentioned before, this work focuses on (mostly) textured images, therefore, for MVTec AD we use the 5 texture classes, each with $\sim 5$ anomaly types. 
MTD contains 1344 images with 5 possible defects, and Leaves includes 422 images covering 4 different kinds of biotic stress.
Similarly with Sohn \etal~\cite{sohn2023anomaly}, we discard the images that contain more than one type of anomaly.
For the Leaves dataset we work on full images and do not perform the specific preprocessing outlined by Esgario \etal~\cite{ESGARIO2020105162}.

\subsection{Metrics}
Normalized mutual information (NMI) is a widely used metric for evaluating clustering results. 
It is calculated as the mutual information between the predicted clustering and the ground truth labels divided by the arithmetic average of their entropies.
We also evaluate the clusters using the adjusted Rand index (ARI)~\cite{hubert1985comparing} and the $F_1$ score~\cite{chinchor1992muc}. 
Since the $F_1$ score requires direct correspondence, we compute the optimal assignment between clusters and labels with the Jonker-Volgenant algorithm~\cite{crouse2016implementing} from SciPy~\cite{SciPy2020}.
For evaluation, we assume the number of clusters to be known; for an extended discussion see Section \ref{sec:purity} in the supplementary.

\subsection{Results}
We compare our system with the state-of-the-art anomaly clustering method of Sohn \etal~\cite{sohn2023anomaly}, including the introduced variants: Average, Maximum Hausdorff, and Weighted Average.
We additionally consider deep image clustering methods (SCAN~\cite{van2020scan}, SPICE~\cite{niu2022spice}) and unsupervised semantic segmentation (STEGO~\cite{hamilton2022unsupervised}) in the comparison. 
Recently, SelFormaly~\cite{lee2023selformaly} introduced a novel approach for solving anomaly detection tasks in a unified manner.
While the authors also present results for anomaly clustering, we note that their method works in a \semisup setting, having a significant advantage compared to our method and the other baselines.

\input figures/qualitative.tex

\Cref{tab:main_results} presents our main results on anomaly clustering; the detailed, per-class scores can be found in the supplementary material (\Cref{sec:breakdown}). 
We obtain a significant improvement in the quality of the clustering over previous methods, agreed by all metrics.
Notably, we outperform SelFormaly~\cite{lee2023selformaly} on MVTec textures despite using a completely unsupervised approach.
Please also see \Cref{tab:cluster_qualitative} for a qualitative assessment.
The improvement brought by our system comes from two sources: high-fidelity blind anomaly localization (BAL) and fine-tuned features obtained from contrastive learning.
We will analyze these contributions in turn in the following ablative study.

\myparagraph{BAL.}
We compare the proposed combination of FCA and VAE residuals with plain zero-shot FCA~\cite{ardelean2024high}. 
The method of Sohn \etal~\cite{sohn2023anomaly} represents the main baseline, using the alpha values before the exponential activation (cf. Equation 5) as anomaly scores. 
On MVTec~AD textures we also compare to a recent dedicated BAL method~\cite{zhang2023s}; only MVTec results are included as the source code is not currently available. 
To show the limitations of applying a \semisup method on anomaly-contaminated data, we include two prominent methods for anomaly detection: DRAEM~\cite{zavrtanik2021draem} and CFA~\cite{lee2022cfa}.
As this evaluation
considers
a binary anomaly localization, we use the common metrics: PRO~\cite{bergmann2021mvtec} with a false positive rate threshold of 0.3, AUROC\textsubscript{p} (pixel level), and AUROC\textsubscript{i} (image level).
The image-level anomaly score is computed as the maximum over pixel-level scores.

\newcommand{\raiseme}{\raisebox{0.4ex}{\strut}}
\begin{table}[ht]
  \centering%
  \begin{tabular*}{\linewidth}{@{\extracolsep{\fill}} l  c c c}
    \toprule
    
    \emph{\raiseme MVTec~AD textures} & PRO & AUROC\textsubscript{p} & AUROC\textsubscript{i} \\
    \cmidrule{2-4}

    ILTM$^\dag$~\cite{patel2023self} & 79.04 & 85.38 & -- \\
    DRAEM~\cite{zavrtanik2021draem} & 31.12 & 57.08 & 73.04 \\
    CFA~\cite{lee2022cfa} & 92.51	& 97.27	& 88.54 \\
    Sohn \etal~\cite{sohn2023anomaly} & 90.84 & 96.61 & 98.69 \\
    Zhang \etal~\cite{zhang2023s} & 91.26 & 96.80 & 97.58 \\
    FCA~\cite{ardelean2024high} & 96.92 & 98.74 & 99.85 \\
    FCA+VAE (ours) & \textbf{97.50} & \textbf{99.02} & \textbf{99.93} \\

    \midrule

    \emph{\raiseme MTD} & PRO & AUROC\textsubscript{p} & AUROC\textsubscript{i} \\
    \cmidrule{2-4}

    DRAEM~\cite{zavrtanik2021draem} & 22.80 & 54.22 & 54.98 \\
    CFA~\cite{lee2022cfa}	& 65.27	& 73.86 & 62.49 \\
    Sohn \etal~\cite{sohn2023anomaly} & 66.51 & 72.86 & 77.00 \\
    FCA~\cite{ardelean2024high} & 72.15 & 74.48 & 80.29 \\
    FCA+VAE (ours) & \textbf{75.53} & \textbf{75.32} & \textbf{83.87} \\

    \midrule

    \emph{\raiseme Leaves} & PRO & AUROC\textsubscript{p} & AUROC\textsubscript{i} \\
    \cmidrule{2-4}

    DRAEM~\cite{zavrtanik2021draem} & 33.53 & 76.91 & 51.34 \\
    CFA~\cite{lee2022cfa} & 76.92 & 97.15 & 81.51 \\
    Sohn \etal~\cite{sohn2023anomaly} & 75.25 & 97.15 & \textbf{92.02} \\
    FCA~\cite{ardelean2024high} & 48.10 & 88.50 & 62.05 \\
    FCA+VAE (ours) & \textbf{77.20} & \textbf{97.62} & 90.62 \\

    \bottomrule

  \end{tabular*}
  \caption{\label{tab:bal_results}%
    Binary anomaly detection and localization metrics in a
    \emph{blind} setting (mixed normal and anomalous data with no
    annotations). For all metrics a higher value is better.\\[0.5ex]
    \parbox{\linewidth}{\hrule width 3em\vspace*{0.5ex}
      \footnotesize{~$^\dag$ ILTM~\cite{patel2023self} results differ slightly in the evaluation protocol.}}
  }
\end{table}

\begin{figure}[ht]
  \setlength{\tabcolsep}{1pt}%
  \renewcommand{\arraystretch}{0.7}%
  \newlength{\imw}%
  \setlength{\imw}{0.29\linewidth}%
  \newcommand{\vlabel}[1]{%
    \raisebox{0.5\imw}{\rotatebox{90}{\clap{\footnotesize\textbf{#1}}}}}%
  \centering%
  \begin{tabular}{@{}c@{\hspace{1mm}}c@{\hspace{1mm}}c@{\hspace{1mm}}c@{\hspace{1mm}}c@{\hspace{1mm}}c@{}}

    \vlabel{Input Image} &
    \frame{\includegraphics[width=\imw]{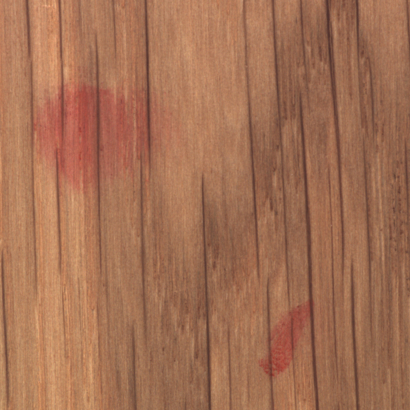}} &
    \frame{\includegraphics[width=\imw]{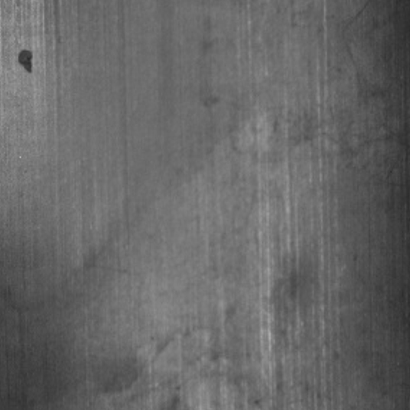}} &
    \frame{\includegraphics[width=\imw]{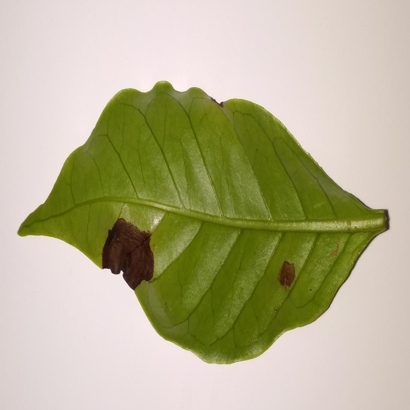}}\\

    \vlabel{GT mask} &
    \frame{\includegraphics[width=\imw]{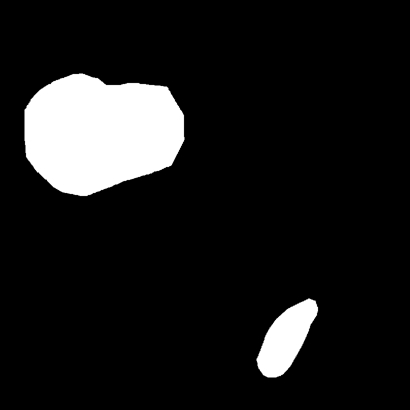}} &
    \frame{\includegraphics[width=\imw]{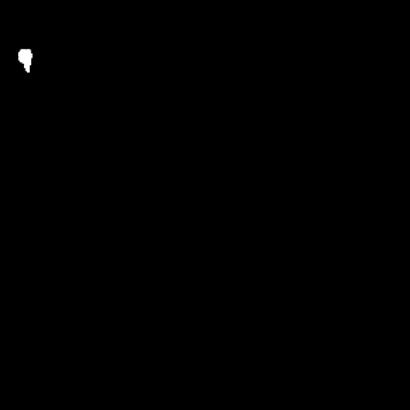}} &
    \frame{\includegraphics[width=\imw]{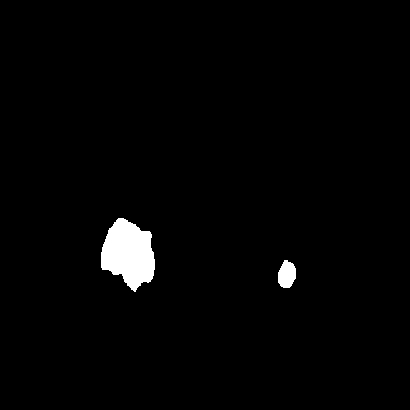}}\\

    \vlabel{Ours} &
    \frame{\includegraphics[width=\imw]{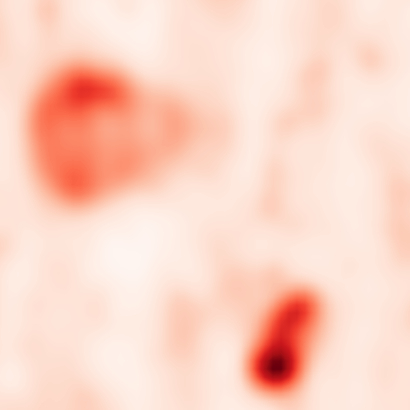}} &
    \frame{\includegraphics[width=\imw]{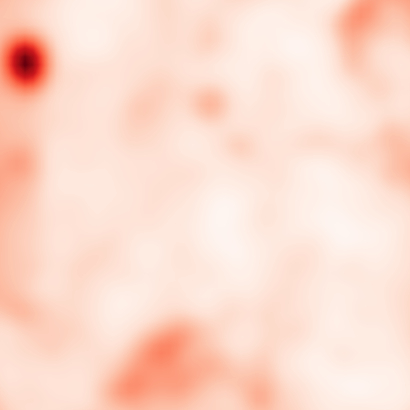}} &
    \frame{\includegraphics[width=\imw]{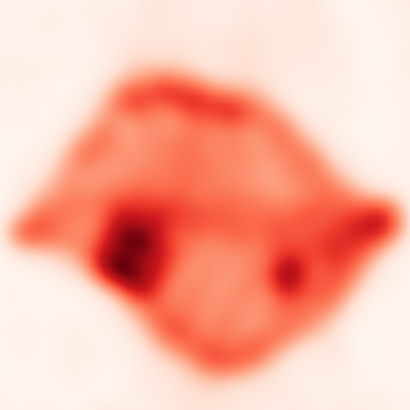}}\\

    \vlabel{FCA} &
    \frame{\includegraphics[width=\imw]{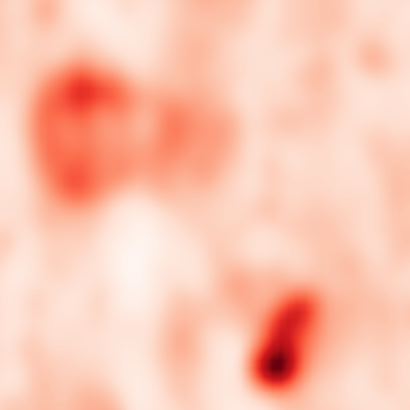}} &
    \frame{\includegraphics[width=\imw]{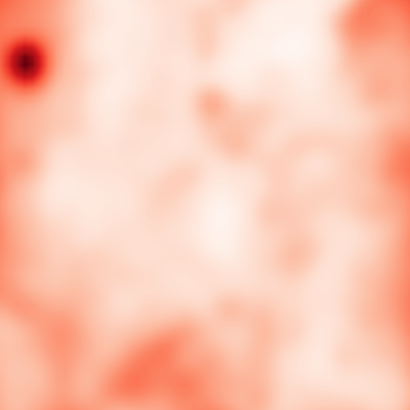}} &
    \frame{\includegraphics[width=\imw]{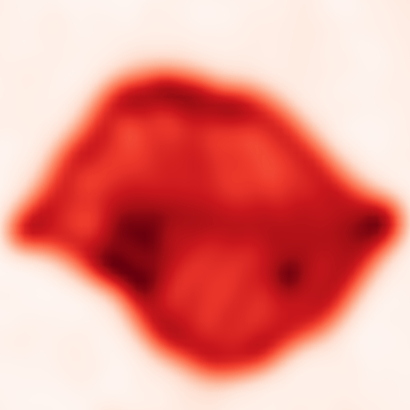}}\\

    \vlabel{Sohn~\etal} &
    \frame{\includegraphics[width=\imw]{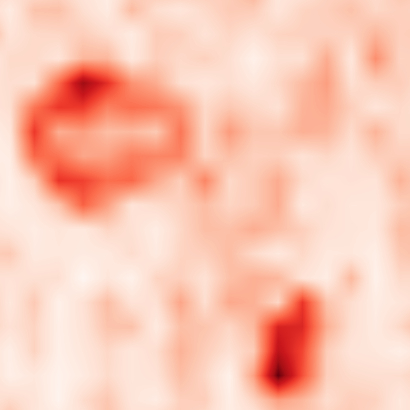}} &
    \frame{\includegraphics[width=\imw]{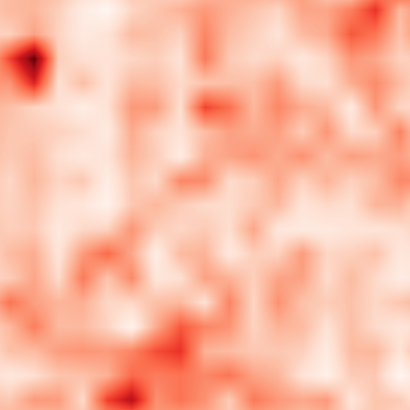}} &
    \frame{\includegraphics[width=\imw]{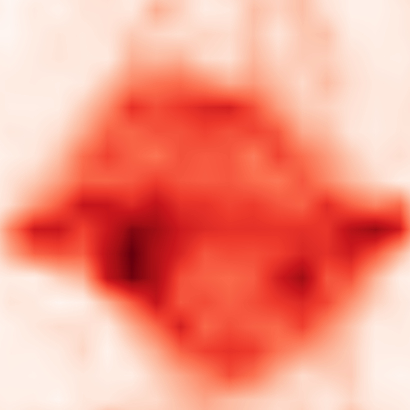}} \\
  \end{tabular}
  \caption{\label{fig:bal_qualitative}%
    Qualitative comparison of anomaly maps produced by different
    approaches to BAL. The images' borders are cropped out.
    \vspace*{-0.75ex}
  }
\end{figure}

The results in Table \ref{tab:bal_results} suggest that FCA performs better than the soft weights from Sohn \etal on mostly homogeneous textures (MVTec, MTD) but the performance degrades on more complex classes (Leaves). 
Our approach improves upon FCA in all cases and surpasses the baseline~\cite{sohn2023anomaly} by a large margin, especially on the size-sensitive localization metric PRO~\cite{bergmann2021mvtec}. 
We also demonstrate the benefit of our approach qualitatively in \Cref{fig:bal_qualitative}.
Our results show better localization through a reduced number of false positives.

\myparagraph{Improved descriptors.}
Intuitively, a finer anomaly localization should improve the downstream performance of the clustering, since it depends on the image-level descriptors (\Cref{eq:descriptor}), computed using the anomaly maps $A_i$.
We verify this by ablating both the VAE and our contrastive learning (CL).
Ablating CL amounts to using the descriptors $D_i$ directly as instance embeddings, similarly to the weighted aggregation from Sohn \etal~\cite{sohn2023anomaly}.
VAE ablation is performed by applying contrastive learning using the anomaly scores obtained from zero-shot FCA directly.

\begin{table*}[ht!]
  \noindent%
  \small\setlength{\tabcolsep}{0.2em}
  \newlength{\csl}\setlength{\csl}{0.35em}
  \newlength{\msl}\setlength{\msl}{0.4em}
  \newcommand{\cs}{\hspace*{\csl}}%
  \newcommand{\ms}{\hspace*{\msl}}%
  \begin{minipage}{0.64\linewidth}
    \centering%
    \begin{tabular}{llccccccccccc}
      \toprule
      \multicolumn{1}{l}{Method} & &
      \multicolumn{3}{c}{MVTec~AD textures} & &
      \multicolumn{3}{c}{MTD} & &
      \multicolumn{3}{c}{Leaves}\\[0.5ex]

      & ~~ &
      NMI & ARI & $F_1$ & ~~ &
      NMI & ARI & $F_1$ & ~~ &
      NMI & ARI & $F_1$ \\

      \cmidrule{3-5}%
      \cmidrule{7-9}%
      \cmidrule{11-13}

      Ours w/o VAE, CL &
      & 0.756 & 0.682 & 0.781 &
      & 0.176 & 0.166 & 0.507 &
      & 0.333 & 0.235 & 0.443 \\
      
      Ours w/o VAE &
      & \textbf{0.807} & \textbf{0.733} & \textbf{0.811} &
      & 0.167 & 0.214 & 0.539 &
      & 0.528 & 0.507 & 0.706 \\ 
      
      Ours w/o CL &
      & 0.769 & 0.693 & 0.775 &
      & 0.142 & 0.161 & 0.513 &
      & 0.512 & 0.461 & 0.588 \\ 
      
      Ours &
      & 0.790 & 0.716 & 0.806 &
      & \textbf{0.221} & \textbf{0.392} & \textbf{0.670} &
      & \textbf{0.712} & \textbf{0.736} & \textbf{0.829} \\
      
      \bottomrule
    \end{tabular}\\[-0.5ex]
    \caption{\label{tab:ablation_clustering}
      Ablation study of the improved anomaly localization (VAE) and contrastive learning (CL). Details in the main text.
    }
  \end{minipage}%
  \hfill%
  \begin{minipage}{0.33\linewidth}
    \centering%
    \begin{tabular}{lccc}
      \toprule
      NMI & MVTec & MTD & Leaves \\
      \midrule

      \kmeans & 0.778 & 0.211 & 0.689 \\
      Agglomerative-Ward & \textbf{0.790} & \textbf{0.221} & 0.712 \\
      Gaussian Mixture & 0.750 & 0.157 & \textbf{0.724} \\
      Spectral & 0.757 & 0.189 & 0.629 \\

      \bottomrule

    \end{tabular}
    \caption{\label{tab:final_clustering}%
      Comparison in terms of NMI of different feature-clustering methods
      applied to the descriptors produced by our method.  }
  \end{minipage}
\end{table*}
\begin{figure*}[ht!]
    \centering
    \begin{subfigure}{0.49\linewidth}
    \includegraphics[width=0.99\linewidth]{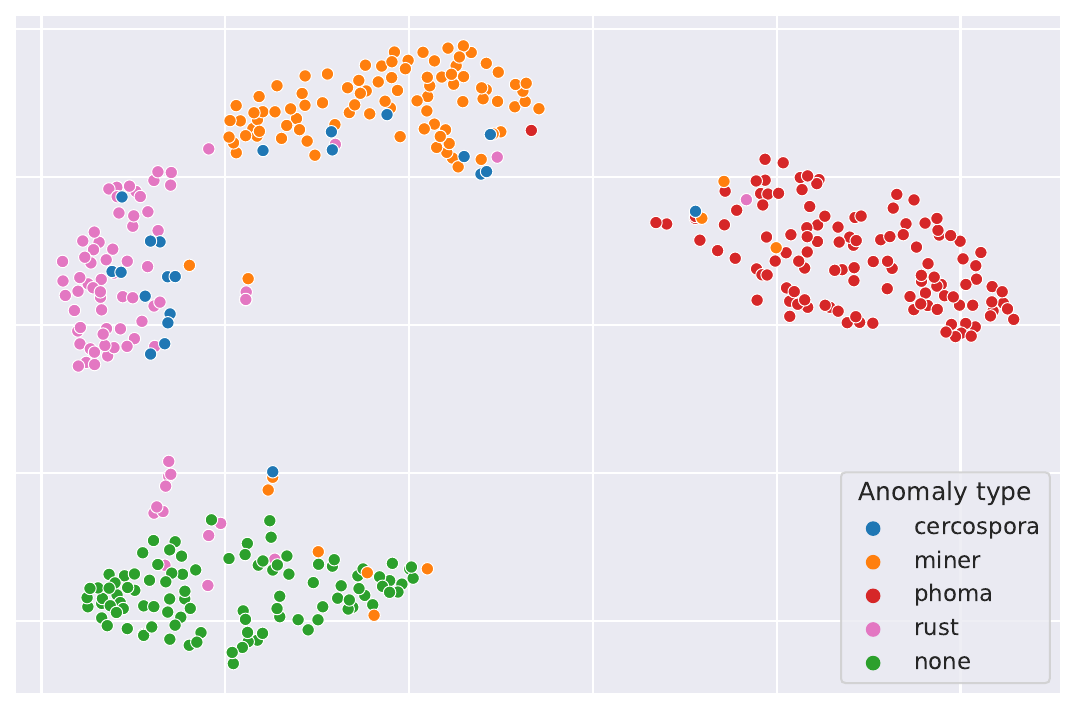}
    \subcaption{Descriptors with contrastive learning}
    \end{subfigure}
  \hfill
  \begin{subfigure}{0.49\linewidth}
    \includegraphics[width=0.99\linewidth]{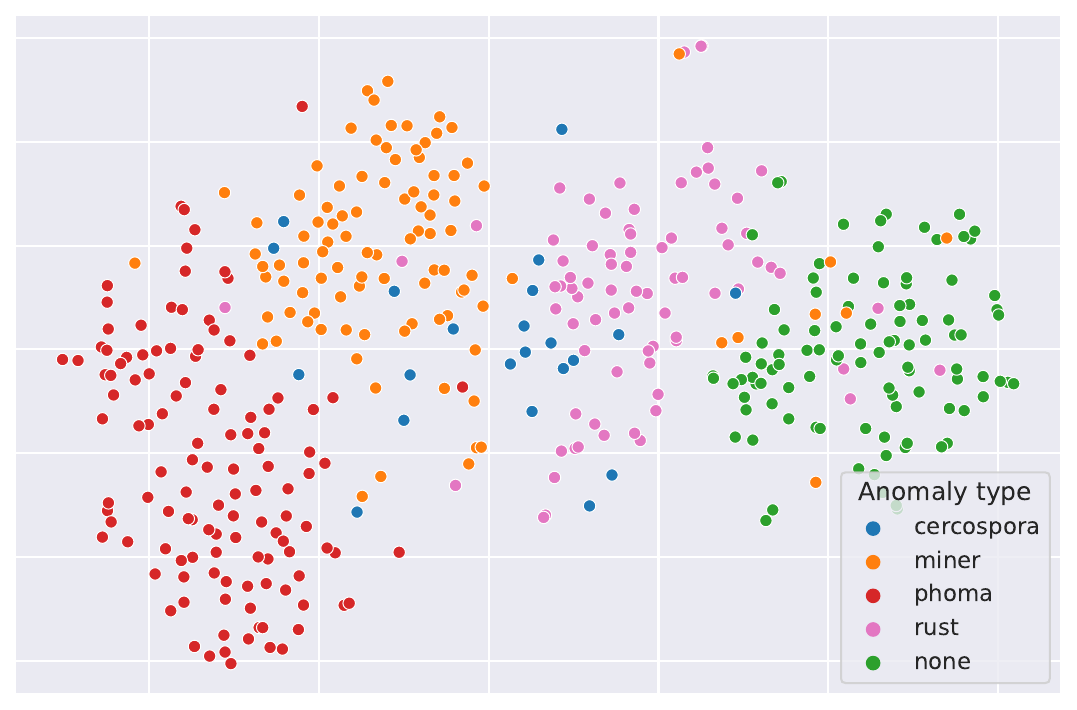}
    \subcaption{Descriptors without contrastive learning}
  \end{subfigure}\\%[-1ex]
  \caption{\label{fig:anomaly_separation}%
    Visualization of the initial image-level descriptors compared to the descriptors computed after contrastive learning. The vectors are projected to two
    dimensions using t-SNE.}
\end{figure*}
\begin{figure*}[ht!]
  \noindent%
  \includegraphics[width=0.33\linewidth]{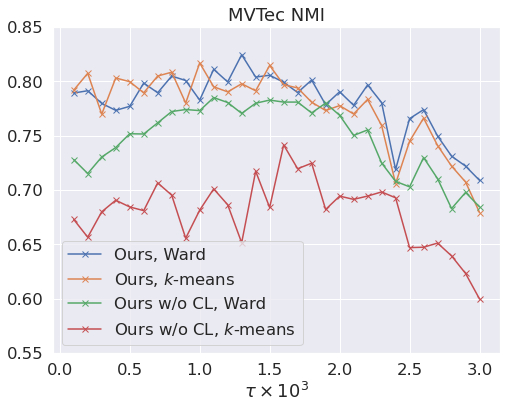}%
  \hfill%
  \includegraphics[width=0.33\linewidth]{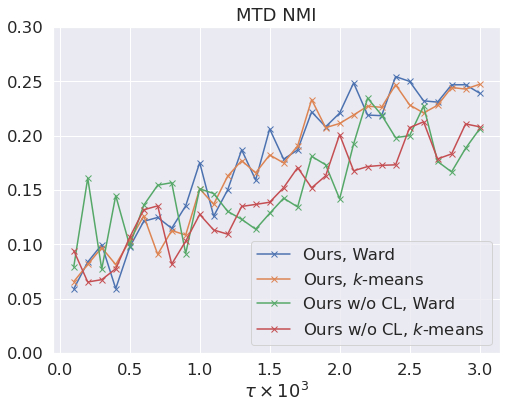}%
  \hfill%
  \includegraphics[width=0.33\linewidth]{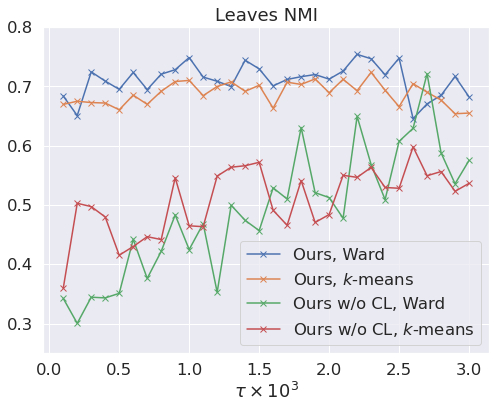}%
  \hspace*{-0.1em}%
  \\[-1.0\baselineskip]
  \caption{\label{fig:sensitivity}%
    Analysis of clustering performance in terms of NMI for varying values of $\tau$. Contrastive learning improves stability.
    \vspace*{-0.4ex}
    }
\end{figure*}

\Cref{tab:ablation_clustering} shows that both components contribute to the performance of our approach. 
As the textures in MVTec~AD are more homogeneous, the benefit of our VAE-based BAL is smaller, and it does not improve the downstream clustering performance on this dataset. Nevertheless, its significance is clearly observed in the results on MTD and Leaves.
To show that our improvements are orthogonal to the final feature-clustering method used, we also present this ablation in the supplementary material (\Cref{tab:ablation_clustering_km}) when using \kmeans instead of Ward clustering.

We visualize the effect of our anomaly-targeted contrastive learning in \Cref{fig:anomaly_separation}.

\myparagraph{Feature-clustering methods.}
The last step of our algorithm consists in an off-the-shelf clustering method based on image-level descriptors.
It can be observed in \Cref{tab:final_clustering} that agglomerative/hierarchical clustering with Ward linkage performs best, with \kmeans being a close second.
Generally, the results obtained with Ward clustering are better compared to \kmeans over different methods~\cite{sohn2023anomaly} and parameters. 
However, as we show in \Cref{sub:sensitivity}, the variance is also larger, with small differences in features yielding considerable different clusterings.
Ideally, the features optimized by our method should make the different anomaly types more easily separable so that the clustering algorithm used would have limited influence on the results. 
Indeed, in \Cref{fig:sensitivity} we show that our contrastive learning has the added benefit of making the algorithm more stable.

\myparagraph{Speed assessment}
Despite being composed of several components and requiring two stages of neural network training, our method is relatively fast. 
Concretely, for the MTD dataset which contains 1344 images, the entire process takes a total of 14 minutes ($7.5\%$ for VAE training, $66.5\%$ for contrastive learning, and $26\%$ for FCA, image-level descriptors calculations and agglomerative clustering).
For reference, the method of Sohn \etal~\cite{sohn2023anomaly} takes 30 minutes due to a time complexity quadratic in the number of images.

\myparagraph{Application domain}
We design our approach specifically for textures, a significant category which occurs often in civil-structures or industrial inspection. For an extended discussion on the scope of our method see \ref{sec:domain} in the supplementary.

\subsection{Sensitivity Analysis}
\label{sub:sensitivity}

We analyze the performance of our algorithm under different values of $\tau$ (see \Cref{eq:descriptor}), the single most important parameter of the method. 
A small value of $\tau$ focuses the weights on a single element in the feature maps, the one with the highest anomaly score.
On the other hand, a large $\tau$ will push the weights toward a uniform distribution, minimizing the effect of using anomaly scores to focus on the regions of interest.
\Cref{fig:sensitivity} shows that different datasets have different optimal values for $\tau$. 
Nevertheless, the range $1.5\cdot 10^{-3}$ to $3.0\cdot 10^{-3}$ seems broadly acceptable for all datasets.

We note here that the improvement added by contrastive learning is visible for various values of $\tau$. Furthermore, the variance of the NMI metric is greatly reduced, as well as the gap between agglomerative (Ward) clustering and \kmeans, which speaks to the stability and robustness of our algorithm.

%% file: figures/qualitative.tex
\begin{figure*}
\setlength{\tabcolsep}{1pt}%
\renewcommand{\arraystretch}{0.3}%
\newlength{\wl}%
\setlength{\wl}{0.04444\textwidth}%
\newcommand{\qualcomplabel}[1]{%
  \footnotesize\textbf{#1}%
  }
\newcommand{\qualcomplabelrot}[1]{%
  \raisebox{0.5\wl}{\rotatebox{90}{\clap{\qualcomplabel{#1}}}}%
  }
\newcommand{\qualcompquad}[4]{%
  \frame{\includegraphics[width=\wl]{images/qualitative/#1}}\hspace*{0.3ex}%
  \frame{\includegraphics[width=\wl]{images/qualitative/#2}}\hspace*{0.3ex}%
  \frame{\includegraphics[width=\wl]{images/qualitative/#3}}\hspace*{0.35ex}%
  \frame{\includegraphics[width=\wl]{images/qualitative/#4}}%
  }
\newcommand{\qualcomprow}[9]{%
  \qualcomplabelrot{\tiny #1}
  & \qualcompquad{#2}{#3}{#4}{#5}
  & \qualcompquad{#6}{#7}{#8}{#9}
  \qualcomprowCont%
}%
\newcommand{\qualcomprowCont}[8]{%
  & \qualcompquad{#1}{#2}{#3}{#4}
  & \qualcompquad{#5}{#6}{#7}{#8}
  \qualcomprowContCont%
}%
\newcommand{\qualcomprowContCont}[4]{%
  & \qualcompquad{#1}{#2}{#3}{#4}
  \\
}%
\newcommand{\beforeOursAndGT}{\multirow{2}{*}{\qualcomplabelrot{Method~~\hspace*{2\wl}}} & }
\newcommand{\betweenOursAndGT}{\\[-0.7ex] & }
\newcommand{\afterOursAndGT}{\\[-0.4444ex]}

\definecolor{cg1}{RGB}{26, 201, 56}
\definecolor{cg2}{RGB}{2, 62, 255}
\definecolor{cg3}{RGB}{255, 124, 0}
\definecolor{cg4}{RGB}{232, 0, 11}
\definecolor{cg5}{RGB}{241, 76, 193}
\definecolor{cg6}{RGB}{0, 215, 255}

\centering
\begin{tabular}{@{}c@{}c@{\hspace{0mm}}c@{\hspace{1mm}}c@{\hspace{1mm}}c@{\hspace{1mm}}c@{\hspace{1mm}}c@{}}
  &
  \qualcomplabel{\qquad} &
  \qualcomplabel{\textcolor{cg2}{Crack}} &
  \qualcomplabel{\textcolor{cg3}{Glue Strip}} &
  \qualcomplabel{\textcolor{cg4}{Gray Stroke}} &
  \qualcomplabel{\textcolor{cg5}{Oil}} &
  \qualcomplabel{\textcolor{cg6}{Rough}} \\[0.75ex]
  \input figures/tile.inc
\end{tabular}\\[0.25\baselineskip]

\begin{tabular}{@{}c@{}c@{\hspace{0mm}}c@{\hspace{1mm}}c@{\hspace{1mm}}c@{\hspace{1mm}}c@{\hspace{1mm}}c@{}}
  &
  \qualcomplabel{\qquad} &
  \qualcomplabel{\textcolor{cg2}{Color}} &
  \qualcomplabel{\textcolor{cg3}{Cut}} &
  \qualcomplabel{\textcolor{cg4}{Fold}} &
  \qualcomplabel{\textcolor{cg5}{Glue}} &
  \qualcomplabel{\textcolor{cg6}{Poke}} \\[0.75ex]
  \input figures/leather.inc
\end{tabular}\\[0.25\baselineskip]

\begin{tabular}{@{}c@{}c@{\hspace{0mm}}c@{\hspace{1mm}}c@{\hspace{1mm}}c@{\hspace{1mm}}c@{\hspace{1mm}}c@{}}
  &
  \qualcomplabel{\qquad} &
  \qualcomplabel{\textcolor{cg2}{Cercospora}} &
  \qualcomplabel{\textcolor{cg3}{Miner}} &
  \qualcomplabel{\textcolor{cg4}{Phoma}} &
  \qualcomplabel{\textcolor{cg5}{Rust}} &
  \qualcomplabel{\textcolor{cg1}{Normal}} \\[0.75ex]
  \input figures/leaves.inc
\end{tabular}\\[-0.5\baselineskip]

\caption{\label{tab:cluster_qualitative}%
    Qualitative evaluation of blind localization and clustering of anomalies. 
    Note that we only show a subset of each cluster; the proportion of misclassifications reflects the true accuracy (rounded down to include more failure cases).
    The color of a column name indicates the contour color associated with that anomaly type; green is reserved for the normal class (no anomaly).
    The contours for `Ours' and `(Ours) w/o CL' differ only through their color because they start from the same binary anomaly localization (our BAL).
    \vspace*{-0.15em}
  }
\end{figure*}

%% file: figures/tile.inc
  \beforeOursAndGT
  \qualcomprow{WA\,\cite{sohn2023anomaly}}
    {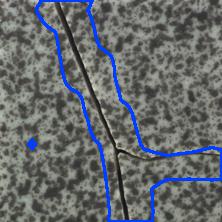}
    {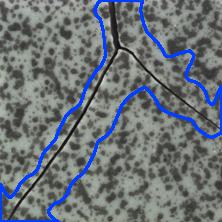}
    {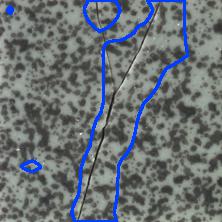}
    {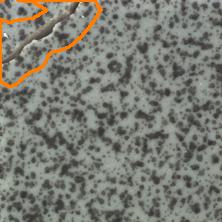}
    {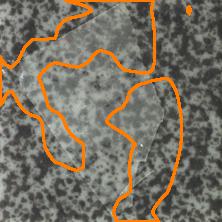}
    {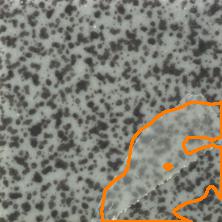} 
    {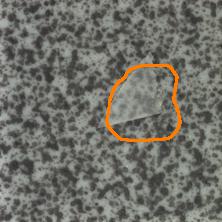}
    {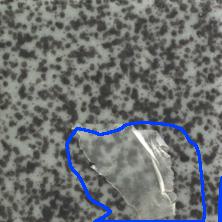} 
    {tile/milwa_gray_stroke_003_gray_stroke.jpg}
    {tile/milwa_gray_stroke_007_gray_stroke.jpg}
    {tile/milwa_gray_stroke_011_gray_stroke.jpg}
    {tile/milwa_gray_stroke_012_good.jpg}
    {tile/milwa_oil_006_oil.jpg}
    {tile/milwa_oil_007_oil.jpg} 
    {tile/milwa_oil_008_oil.jpg} 
    {tile/milwa_oil_009_oil.jpg}
    {tile/milwa_rough_007_rough.jpg}
    {tile/milwa_rough_011_rough.jpg}
    {tile/milwa_rough_000_rough.jpg}
    {tile/milwa_rough_002_good.jpg}
  \betweenOursAndGT
  \qualcomprow{w/o CL}
    {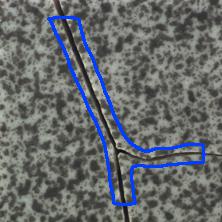}
    {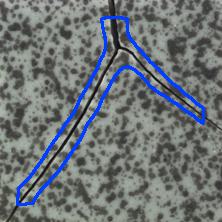}
    {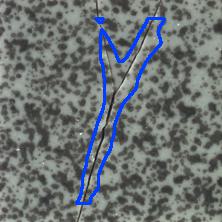}
    {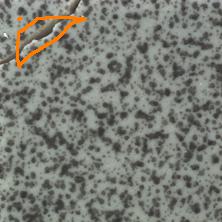}
    {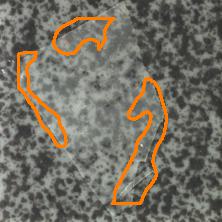}
    {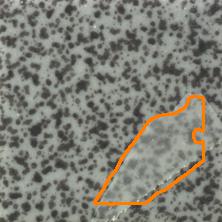}
    {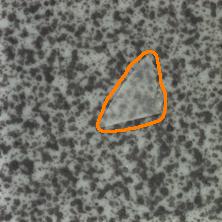}
    {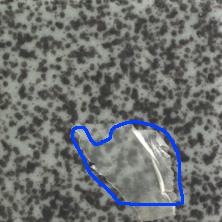}
    {tile/nocl_gray_stroke_003_gray_stroke.jpg}
    {tile/nocl_gray_stroke_007_gray_stroke.jpg}
    {tile/nocl_gray_stroke_011_gray_stroke.jpg}
    {tile/nocl_gray_stroke_012_good.jpg}
    {tile/nocl_oil_006_oil.jpg}
    {tile/nocl_oil_007_oil.jpg}
    {tile/nocl_oil_008_oil.jpg}
    {tile/nocl_oil_009_oil.jpg}
    {tile/nocl_rough_007_rough.jpg}
    {tile/nocl_rough_011_rough.jpg}
    {tile/nocl_rough_000_gray_stroke.jpg}
    {tile/nocl_rough_002_gray_stroke.jpg}
  \betweenOursAndGT
  \qualcomprow{Ours}
    {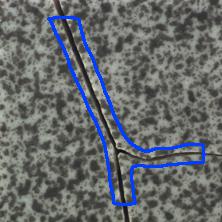}
    {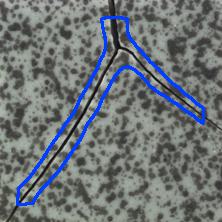}
    {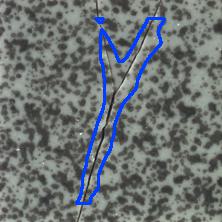}
    {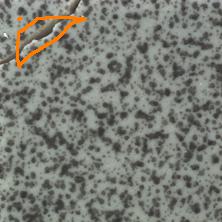}
    {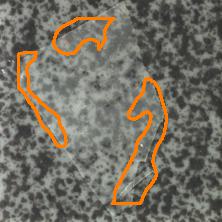}
    {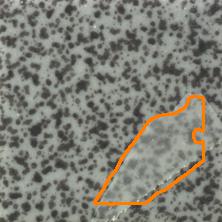}
    {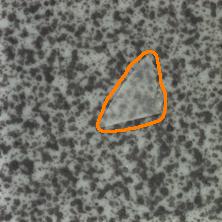}
    {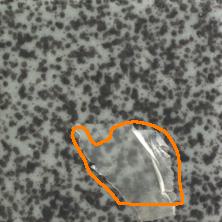}
    {tile/ours_gray_stroke_003_gray_stroke.jpg}
    {tile/ours_gray_stroke_007_gray_stroke.jpg}
    {tile/ours_gray_stroke_011_gray_stroke.jpg}
    {tile/ours_gray_stroke_012_good.jpg}
    {tile/ours_oil_006_oil.jpg}
    {tile/ours_oil_007_oil.jpg}
    {tile/ours_oil_008_oil.jpg}
    {tile/ours_oil_009_oil.jpg}
    {tile/ours_rough_007_rough.jpg}
    {tile/ours_rough_011_rough.jpg}
    {tile/ours_rough_000_rough.jpg}
    {tile/ours_rough_002_rough.jpg}

  \betweenOursAndGT
  \qualcomprow{GT}
    {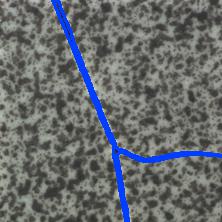}
    {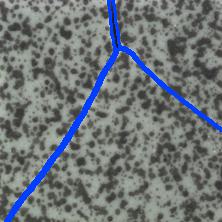}
    {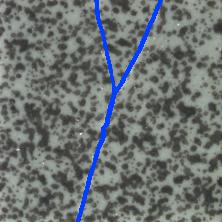}
    {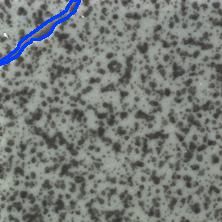}
    {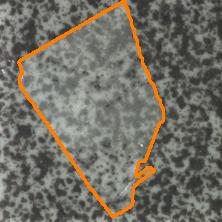}
    {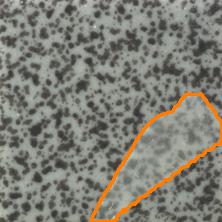}
    {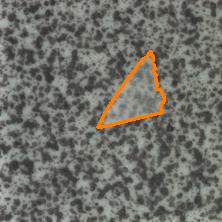}
    {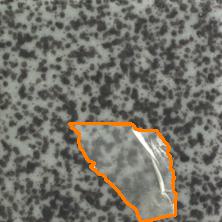}
    {tile/gt_gray_stroke_003.jpg}
    {tile/gt_gray_stroke_007.jpg}
    {tile/gt_gray_stroke_011.jpg}
    {tile/gt_gray_stroke_012.jpg}
    {tile/gt_oil_006.jpg}
    {tile/gt_oil_007.jpg}
    {tile/gt_oil_008.jpg}
    {tile/gt_oil_009.jpg}
    {tile/gt_rough_007.jpg}
    {tile/gt_rough_011.jpg}
    {tile/gt_rough_000.jpg}
    {tile/gt_rough_002.jpg}
  \afterOursAndGT

%% file: figures/leather.inc
  \beforeOursAndGT
  \qualcomprow{WA\,\cite{sohn2023anomaly}}
    {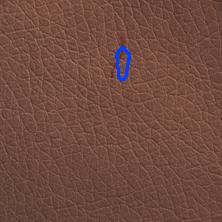}
    {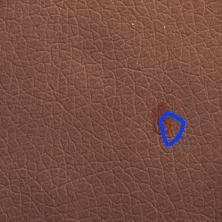}
    {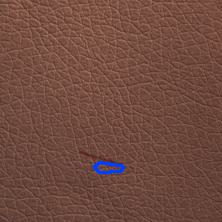}
    {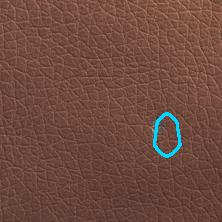}
    {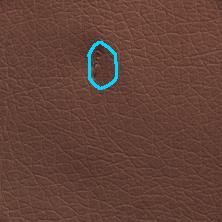}
    {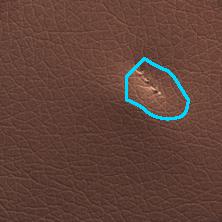}
    {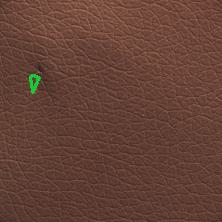}
    {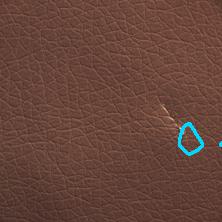}
    {leather/milwa_fold_000_fold.jpg}
    {leather/milwa_fold_016_fold.jpg}
    {leather/milwa_fold_007_poke.jpg}
    {leather/milwa_fold_013_poke.jpg}
    {leather/milwa_glue_017_glue.jpg}
    {leather/milwa_glue_011_poke.jpg}
    {leather/milwa_glue_012_good.jpg}
    {leather/milwa_glue_004_good.jpg}
    {leather/milwa_poke_003_poke.jpg}
    {leather/milwa_poke_000_poke.jpg}
    {leather/milwa_poke_017_poke.jpg}
    {leather/milwa_poke_013_good.jpg}
  \betweenOursAndGT
  \qualcomprow{w/o CL}
    {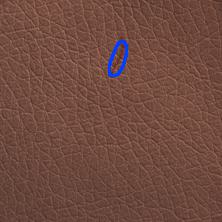}
    {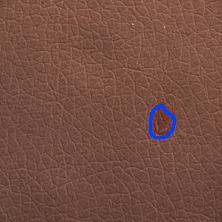}
    {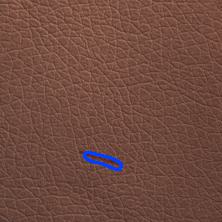}
    {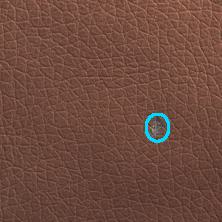}
    {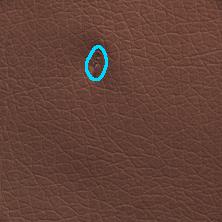}
    {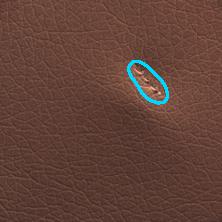}
    {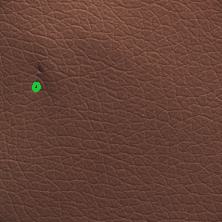}
    {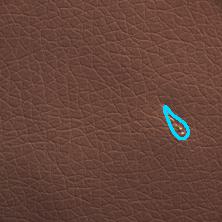}
    {leather/nocl_fold_000_fold.jpg}
    {leather/nocl_fold_016_fold.jpg}
    {leather/nocl_fold_007_poke.jpg}
    {leather/nocl_fold_013_good.jpg}
    {leather/nocl_glue_017_glue.jpg}
    {leather/nocl_glue_011_good.jpg}
    {leather/nocl_glue_012_good.jpg}
    {leather/nocl_glue_004_good.jpg}
    {leather/nocl_poke_003_poke.jpg}
    {leather/nocl_poke_000_poke.jpg}
    {leather/nocl_poke_017_poke.jpg}
    {leather/nocl_poke_013_poke.jpg}
  \betweenOursAndGT
  \qualcomprow{Ours}
    {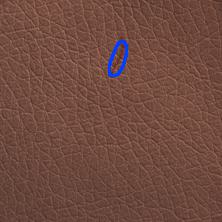}
    {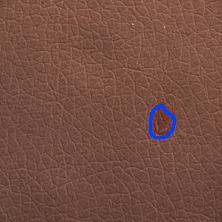}
    {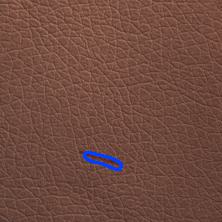}
    {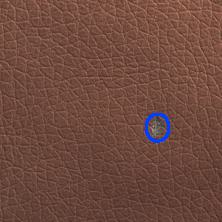}
    {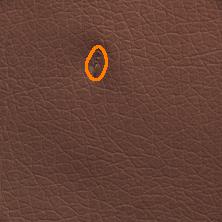}
    {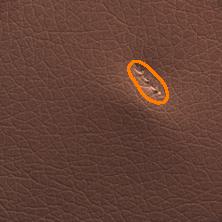}
    {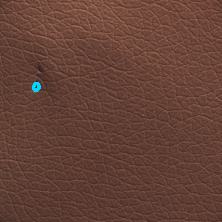}
    {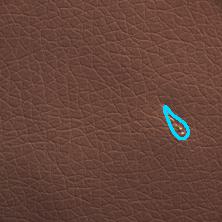}
    {leather/ours_fold_000_fold.jpg}
    {leather/ours_fold_016_fold.jpg}
    {leather/ours_fold_007_fold.jpg}
    {leather/ours_fold_013_poke.jpg}
    {leather/ours_glue_017_glue.jpg}
    {leather/ours_glue_011_glue.jpg}
    {leather/ours_glue_012_glue.jpg}
    {leather/ours_glue_004_good.jpg}
    {leather/ours_poke_003_poke.jpg}
    {leather/ours_poke_000_cut.jpg}
    {leather/ours_poke_017_cut.jpg}
    {leather/ours_poke_013_cut.jpg}

  \betweenOursAndGT
  \qualcomprow{GT}
    {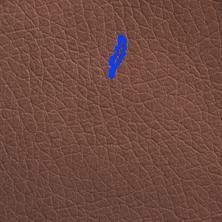}
    {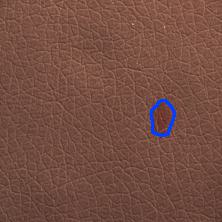}
    {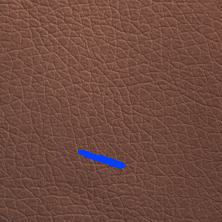}
    {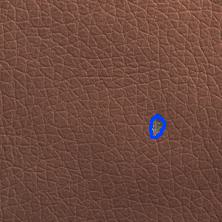}
    {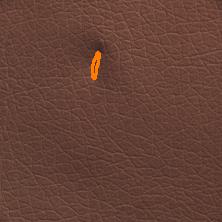}
    {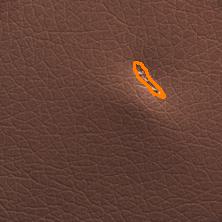}
    {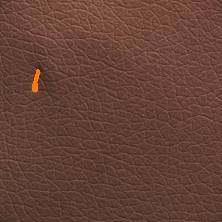}
    {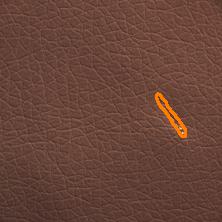}
    {leather/gt_fold_000.jpg}
    {leather/gt_fold_016.jpg}
    {leather/gt_fold_007.jpg}
    {leather/gt_fold_013.jpg}
    {leather/gt_glue_017.jpg}
    {leather/gt_glue_011.jpg}
    {leather/gt_glue_012.jpg}
    {leather/gt_glue_004.jpg}
    {leather/gt_poke_003.jpg}
    {leather/gt_poke_000.jpg}
    {leather/gt_poke_017.jpg}
    {leather/gt_poke_013.jpg}
  \afterOursAndGT

%% file: figures/leaves.inc
  \beforeOursAndGT
  \qualcomprow{WA\,\cite{sohn2023anomaly}}
    {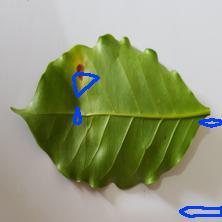}
    {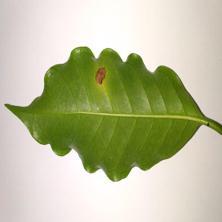}
    {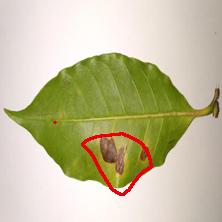}
    {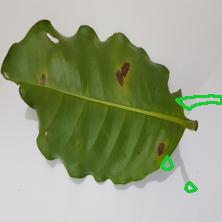}
    {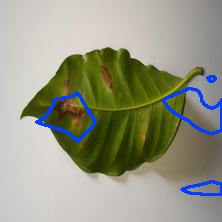}
    {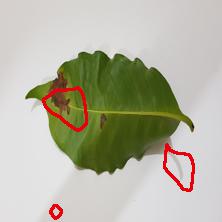}
    {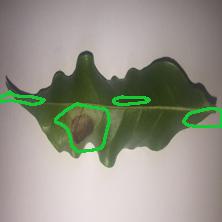}
    {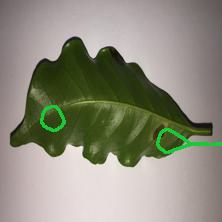}
    {leaves/milwa_phoma_242_phoma.jpg}
    {leaves/milwa_phoma_287_phoma.jpg}
    {leaves/milwa_phoma_360_good.jpg}
    {leaves/milwa_phoma_666_phoma.jpg}
    {leaves/milwa_rust_99_rust.jpg}
    {leaves/milwa_rust_150_rust.jpg}
    {leaves/milwa_rust_880_cercospora.jpg}
    {leaves/milwa_rust_1439_cercospora.jpg}
    {leaves/milwa_good_1134_good.jpg}
    {leaves/milwa_good_693_good.jpg}
    {leaves/milwa_good_766_good.jpg}
    {leaves/milwa_good_446_cercospora.jpg}

  \betweenOursAndGT
  \qualcomprow{w/o CL}
    {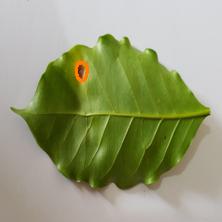}
    {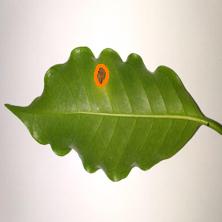}
    {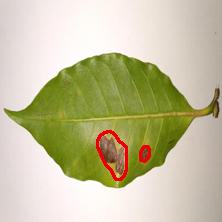}
    {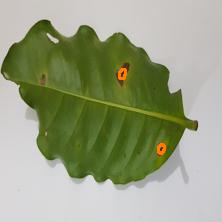}
    {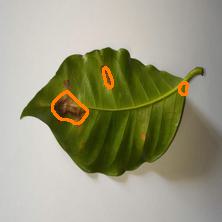}
    {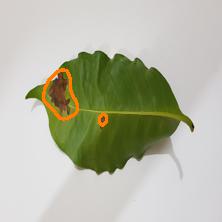}
    {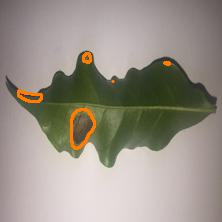}
    {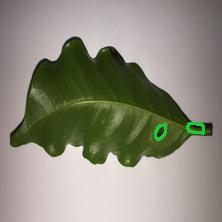}
    {leaves/nocl_phoma_242_rust.jpg}
    {leaves/nocl_phoma_287_rust.jpg}
    {leaves/nocl_phoma_360_rust.jpg}
    {leaves/nocl_phoma_666_rust.jpg}
    {leaves/nocl_rust_99_miner.jpg}
    {leaves/nocl_rust_150_miner.jpg}
    {leaves/nocl_rust_880_miner.jpg}
    {leaves/nocl_rust_1439_miner.jpg}
    {leaves/nocl_good_1134_good.jpg}
    {leaves/nocl_good_693_good.jpg}
    {leaves/nocl_good_766_rust.jpg}
    {leaves/nocl_good_446_good.jpg}
  \betweenOursAndGT
  \qualcomprow{Ours}
    {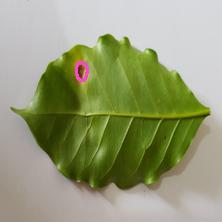}
    {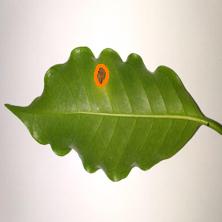}
    {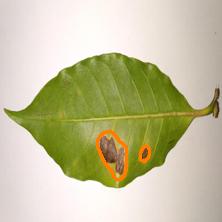}
    {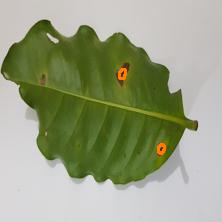}
    {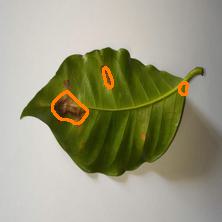}
    {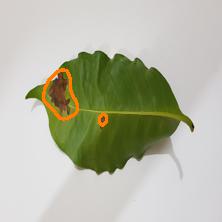}
    {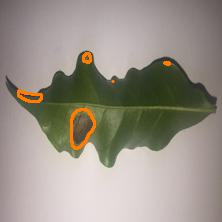}
    {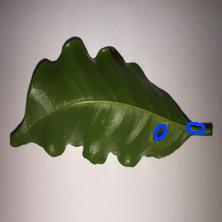}
    {leaves/ours_phoma_242_phoma.jpg}
    {leaves/ours_phoma_287_phoma.jpg}
    {leaves/ours_phoma_360_phoma.jpg}
    {leaves/ours_phoma_666_miner.jpg}
    {leaves/ours_rust_99_rust.jpg}
    {leaves/ours_rust_150_rust.jpg}
    {leaves/ours_rust_880_rust.jpg}
    {leaves/ours_rust_1439_good.jpg}
    {leaves/ours_good_1134_good.jpg}
    {leaves/ours_good_693_good.jpg}
    {leaves/ours_good_766_good.jpg}
    {leaves/ours_good_446_cercospora.jpg}

  \betweenOursAndGT
  \qualcomprow{GT}
    {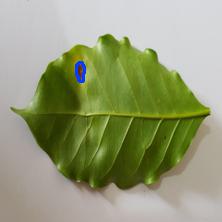}
    {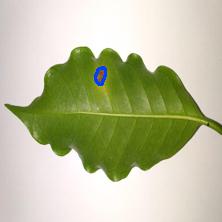}
    {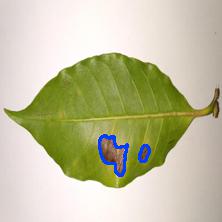}
    {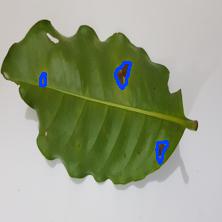}
    {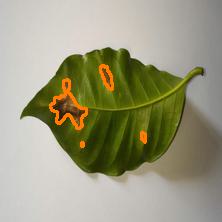}
    {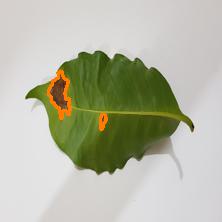}
    {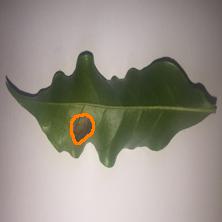}
    {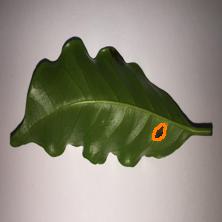}
    {leaves/gt_phoma_242.jpg}
    {leaves/gt_phoma_287.jpg}
    {leaves/gt_phoma_360.jpg}
    {leaves/gt_phoma_666.jpg}
    {leaves/gt_rust_99.jpg}
    {leaves/gt_rust_150.jpg}
    {leaves/gt_rust_880.jpg}
    {leaves/gt_rust_1439.jpg}
    {leaves/gt_good_1134.jpg}
    {leaves/gt_good_693.jpg}
    {leaves/gt_good_766.jpg}
    {leaves/gt_good_446.jpg}
  \afterOursAndGT

%% file: sec/5_conclusion.tex
\section{Conclusion}
\label{sec:conclusion}

In this work, we introduce a novel pipeline for clustering anomalies in textured surfaces.
Firstly, we extend the state-of-the-art zero-shot method of Ardelean and Weyrich~\cite{ardelean2024high} for anomaly localization, and enable incorporating additional information from multiple images. This vastly improves the performance on more complex, non-stationary images, and establishes a new state of the art for the blind anomaly localization task.
Secondly, we leverage the predicted anomaly maps to mine positive and negative pairs and employ contrastive learning. This improves the descriptiveness of the neural features, increasing the separability of the anomaly classes.
We validate the contributions extensively by comparing with prior art on three different datasets, and we include a comprehensive ablation study that clarifies the role of each component of our method.

\myparagraph{Acknowledgements.}
This project has received funding from the European Union’s Horizon 2020 research and innovation programme under the Marie Skłodowska-Curie grant agreement No 956585.

%% file: sec/X_suppl.tex
\clearpage
\setcounter{page}{1}
\maketitlesupplementary

\setcounter{section}{0}
\renewcommand\thesection{S\arabic{section}}

\section{Summary}
This supplementary material includes additional insights regarding the proposed anomaly clustering method in the form of visualizations, comparisons, evaluations, and details useful for the reproducibility of our results. 

\section{Detailed clustering evaluation}
\label{sec:breakdown}

In \Cref{tab:detailed_results}, we present the metrics' breakdown at the level of texture classes for MVTec~AD~\cite{bergmann2021mvtec}. MTD and Leaves datasets are not included since they contain a single texture class only.

\begin{table*}[ht]
  \centering%
  \newcommand{\cs}{\quad\quad}%
    \scalebox{0.77}{
  \begin{tabular}{l@{\cs}ccc@{\cs}ccc@{\cs}ccc@{\cs}ccc@{\cs}ccc}
    \toprule
    \multicolumn{1}{l}{Method} &
    \multicolumn{3}{c@{\cs}}{Carpet} &
    \multicolumn{3}{c@{\cs}}{Grid} &
    \multicolumn{3}{c@{\cs}}{Leather} &
    \multicolumn{3}{c@{\cs}}{Tile} &
    \multicolumn{3}{c}{Wood} \\[0.5ex]

    &
    NMI & ARI & $F_1$ &
    NMI & ARI & $F_1$ &
    NMI & ARI & $F_1$ &
    NMI & ARI & $F_1$ &
    NMI & ARI & $F_1$ \\

    \midrule

    SPICE~\cite{niu2022spice}
    & 0.202 & 0.031 & 0.231
    & 0.197 & 0.033 & 0.166
    & 0.373 & 0.153 & 0.016
    & 0.284 & 0.490 & 0.176 
    & 0.284 & 0.130 & 0.282 \\

    STEGO~\cite{hamilton2022unsupervised}
    & 0.271 & 0.167 & 0.402
    & 0.123 & 0.002 & 0.282
    & 0.338 & 0.202 & 0.395
    & 0.790 & 0.707 & 0.701 
    & 0.423 & 0.274 & 0.515 \\
    
    Average~\cite{sohn2023anomaly}
    & 0.287 & 0.138 & 0.392
    & 0.158 & 0.033 & 0.326
    & 0.398 & 0.218 & 0.465
    & 0.288 & 0.157 & 0.444
    & 0.231 & 0.066 & 0.384 \\
    
    Max. Hausdorff~\cite{sohn2023anomaly}
    & \textbf{0.660} & \textbf{0.586} & \textbf{0.795}
    & 0.129 & 0.018 & 0.308
    & 0.725 & 0.652 & 0.762
    & 0.932 & 0.914 & 0.957
    & 0.678 & 0.500 & 0.716 \\
    
    Weighted Average~\cite{sohn2023anomaly}
    & 0.656 & 0.576 & 0.647
    & 0.143 & 0.018 & 0.304
    & 0.778 & 0.674 & 0.704
    & 0.933 & 0.921 & 0.957
    & \textbf{0.860} & \textbf{0.815} & 0.921 \\

    Ours
    & 0.617 & 0.458 & 0.590
    & \textbf{0.710} & \textbf{0.620} & \textbf{0.718}
    & \textbf{0.832} & \textbf{0.779} & \textbf{0.823}
    & \textbf{0.947} & \textbf{0.936} & \textbf{0.974}
    & 0.846 & 0.788 & \textbf{0.926} \\

    \bottomrule
  \end{tabular}}\\[-0.25\baselineskip]
  \caption{\label{tab:detailed_results}
    Detailed breakdown of the metrics generated by different anomaly clustering methods on the MVTec AD~\cite{bergmann2021mvtec} textures. The results for the algorithms introduced by Sohn \etal are lifted from the original publication~\cite{sohn2023anomaly}.
  }
\end{table*}

\section{Additional ablation results.}
In \Cref{tab:ablation_clustering_km}, we present an ablation study in a similar manner to \Cref{tab:ablation_clustering} from the main text.
The difference is that these results are obtained under \kmeans clustering as the final step instead of agglomerative clustering with Ward linkage.
The results show that our contributions hold independent from a particular choice of feature-clustering method.

\begin{table}[ht]
  \centering%
  \begin{tabular*}{\linewidth}{@{\extracolsep{\fill}} l  c c c}
    \toprule
    
    \emph{\raiseme MVTec~AD textures} & NMI & ARI & $F_1$ \\
    \cmidrule{2-4}

    Ours w/o VAE, CL & 0.667 & 0.589 & 0.745 \\
    Ours w/o VAE & 0.773 & 0.678 & 0.801 \\
    Ours w/o CL & 0.694 & 0.605 & 0.743 \\
    Ours & \textbf{0.778} & \textbf{0.701} & \textbf{0.809} \\

    \midrule

    \emph{\raiseme MTD} & NMI & ARI & $F_1$ \\
    \cmidrule{2-4}

    Ours w/o VAE, CL & 0.148 & 0.109 & 0.411 \\
    Ours w/o VAE & 0.154 & 0.181 & 0.499 \\
    Ours w/o CL & 0.201 & 0.253 & 0.560 \\
    Ours & \textbf{0.211} & \textbf{0.347} & \textbf{0.649} \\

    \midrule

    \emph{\raiseme Leaves} & NMI & ARI & $F_1$ \\
    \cmidrule{2-4}

    Ours w/o VAE, CL & 0.456 & 0.426 & 0.635 \\
    Ours w/o VAE & 0.568 & 0.561 & 0.735 \\
    Ours w/o CL & 0.484 & 0.412 & 0.659 \\
    Ours & \textbf{0.689} & \textbf{0.704} & \textbf{0.801} \\

    \bottomrule

  \end{tabular*}
\caption{\label{tab:ablation_clustering_km}
    Ablation results when using \kmeans for the final feature-level clustering.
}
\end{table}

To develop our blind anomaly localization system, we build on FCA~\cite{ardelean2024high} as a state-of-the-art zero-shot anomaly localization method.
We argue that this zero-shot detection synergizes well with the VAE reconstruction, fully capitalizing on the information in the residual maps (computed according to \Cref{eq:bal}).
In \Cref{tab:ablate_fca}, we substantiate our decision by replacing FCA as the residual maps processor with different ways to compare the original features to the reconstruction.
We compare against various approaches from the relevant literature: $L^1$ and $L^2$ norm~\cite{Baur2018DeepAM, chen2018deep}, SSIM~\cite{Bergmann2018ImprovingUD}, and Rec-grad~\cite{zimmerer2019unsupervised, zhou2020unsupervised}.

\begin{table}
  \centering
  \begin{tabular}{lccc}
    \toprule
    MVTec AD textures & PRO & AUROC\textsubscript{p} & AUROC\textsubscript{i} \\
    \midrule

    VAE + $L^1$ & 95.24 & 97.73 & 98.41 \\
    VAE + $L^2$ & 94.68 & 97.39 & 98.02  \\
    VAE + SSIM  & 96.73 & 98.73 & 99.77 \\
    VAE + Rec-grad  & 82.00 & 90.33 & 94.82 \\
    VAE + FCA (ours) & \textbf{97.50} & \textbf{99.02} & \textbf{99.93} \\
    \bottomrule

  \end{tabular}
  \caption{\label{tab:ablate_fca}%
    Results for FCA ablation. The VAE feature reconstruction yields the best results when combined with FCA as the residual maps processor.
  }
\end{table}

There are, of course, many other levels on which we could analyze our pipeline, including the effect of different hyperparameters.
However, we observe that most parameters of our method, such as the number of layers in the VAE and feature refiner $H$, number of neighbors $k$ and margin for contrastive learning, as well as the optimization parameters, have limited effect on the final performance or may offer benefits only for specific datasets.
In any case, we did not perform an exhaustive hyperparameter search, and a different configuration could potentially yield better results.

\section{Automatic threshold estimation}
\label{sec:threshold}

Our contrastive learning formulation assumes binary anomaly maps. 
In order to maintain a fully unsupervised method, we propose an algorithm to estimate the threshold $t$ for binarization.
One can observe from the blind anomaly localization results (\Cref{tab:bal_results}) that image-level anomaly scores are reasonably reliable (83\% on MTD and almost a perfect result on MVTec AD).
Therefore, if we knew in advance the proportion of normal samples in the input set, we could estimate the threshold as the corresponding quantile in the predicted image-level anomaly scores.

However, in a fully unsupervised scenario, this proportion is also unknown.
To estimate this value, we perform an initial \kmeans clustering using descriptors $D_i$ (see \Cref{eq:descriptor}).
We then find the normality cluster by taking the group with the smallest average anomaly score.
The size of this cluster divided by the total number of images is finally used as the ratio of normal images to compute the threshold $t$.

\section{Unknown number of anomaly types}
\label{sec:purity}

\begin{figure}
    \centering
    \includegraphics[width=\linewidth]{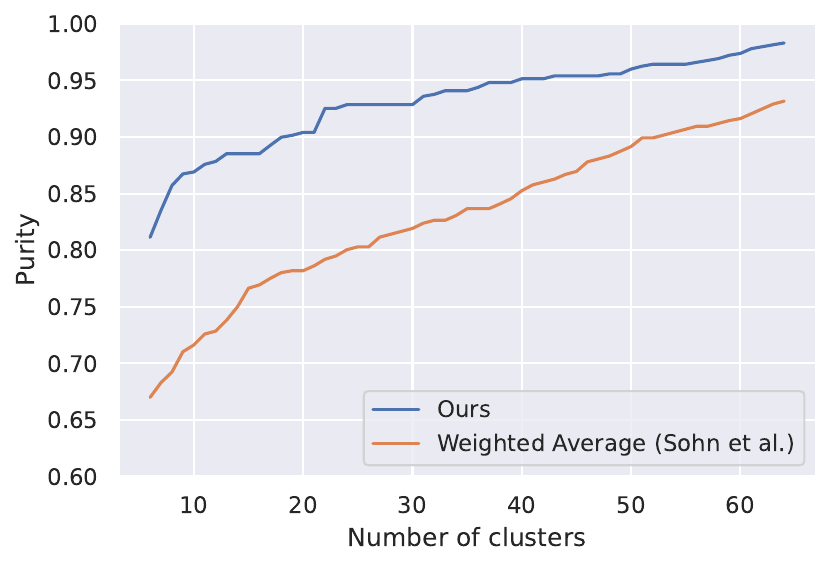}
    \caption{\label{fig:purity} Visualization of the increase of purity with respect to over-clustering. The purity is averaged over all MVTec AD textures. }
    
\end{figure}

The quantitative and qualitative results in the main paper are obtained under the common assumption that the number of anomaly types is known in advance. This is made primarily to facilitate comparison across different methods. However, in practice, the anomaly types residing in the input set might be unknown.

We note that up to the final feature-based clustering, our algorithm is not constrained to a predefined number of classes\footnote{This excludes the heuristic for finding the threshold $t$; see \Cref{sec:threshold}}.
This offers great flexibility as the image descriptors can be used in other ways such as the automatic discovery of the number of clusters or as a tool to reduce manual labeling efforts (through over-clustering). 
We show this in \Cref{fig:purity} by plotting the purity as a function of the number of clusters.

\section{Qualitative pixel-level results.}
\begin{figure}[ht]
  \setlength{\tabcolsep}{1pt}%
  \renewcommand{\arraystretch}{0.7}%
  \setlength{\imw}{0.31\linewidth}%
  \newcommand{\vlabel}[1]{%
    \raisebox{0.5\imw}{\rotatebox{90}{\clap{\footnotesize\textbf{#1}}}}}%
  \centering%
  \begin{tabular}{@{}c@{\hspace{1mm}}c@{\hspace{1mm}}c@{\hspace{1mm}}c@{\hspace{1mm}}c@{\hspace{1mm}}c@{}}

    \vlabel{MVTec wood} &
    \frame{\includegraphics[width=\imw]{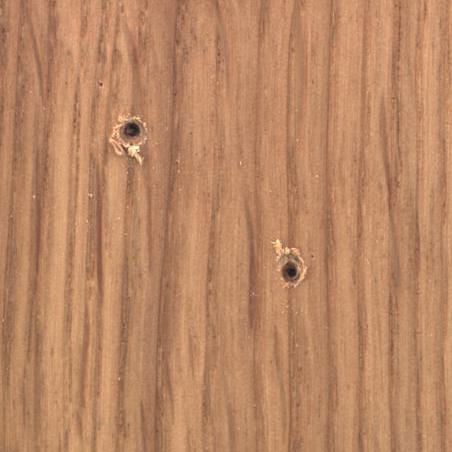}} &
    \frame{\includegraphics[width=\imw]{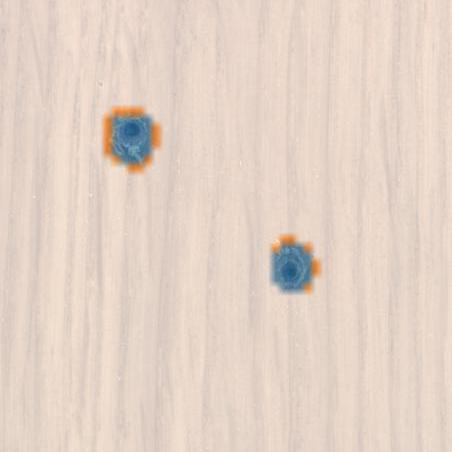}} &
    \frame{\includegraphics[width=\imw]{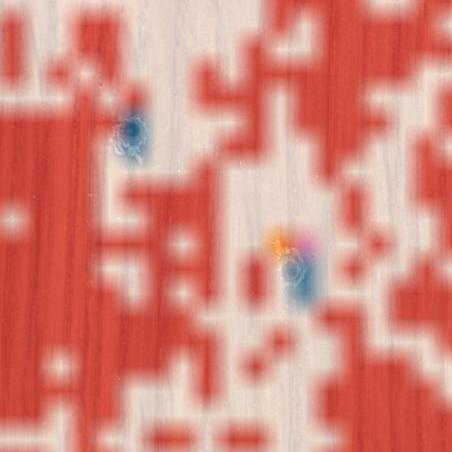}}\\

    \vlabel{Leaves} &
    \frame{\includegraphics[width=\imw]{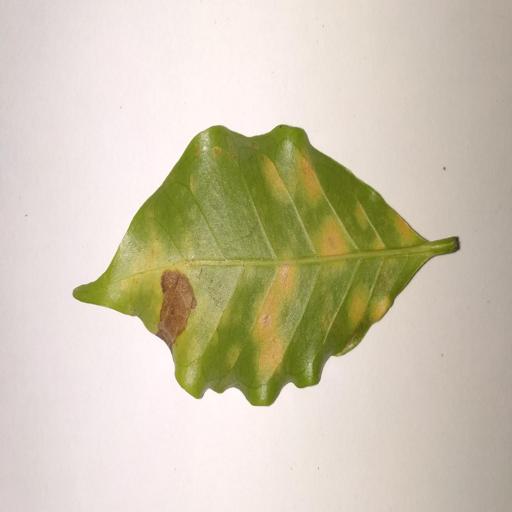}} &
    \frame{\includegraphics[width=\imw]{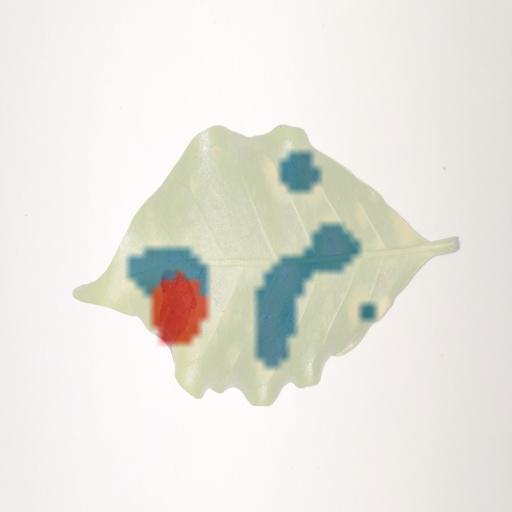}} &
    \frame{\includegraphics[width=\imw]{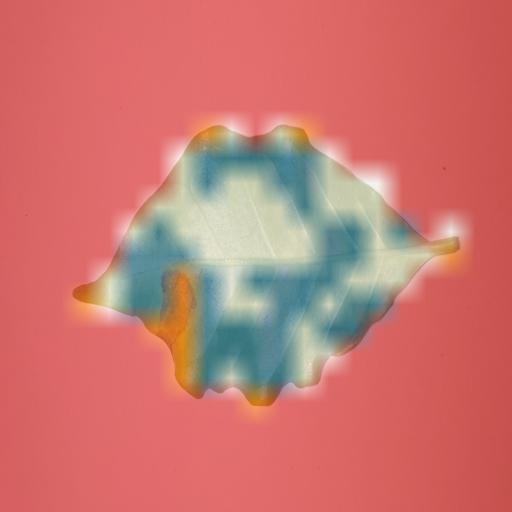}}\\

    & Input image & Ours & Sohn \etal~\cite{sohn2023anomaly}

  \end{tabular}
  \caption{\label{fig:pix_seg}%
    Qualitative comparison for pixel-level clustering on unseen images.
    We depict the predicted anomaly types with the following colors for wood: blue -- hole, orange -- scratch, red -- color stain, and for leaves: blue -- cercospora, orange -- miner, red -- phoma. In both cases white is used for pixels classified as \emph{normal}.
  }
\end{figure}

Our contrastive training enables us to leverage the improved features for each pixel in the input image. In \Cref{fig:pix_seg}, we show that one can use the features to perform clustering at a pixel level, essentially achieving multi-class anomaly segmentation/localization. For this experiment we use unseen images (not used for contrastive learning or training the VAE): an image from the \emph{combined} subcategory of the wood texture in MVTec~AD and a leaf texture that contains two different anomaly types. The wood texture initially had two types of anomalies; however, we cropped the image to include only one type (holes) because the other (knot) was not part of the fitting set (\ie, the subcategories: hole, scratch, color, liquid, and normal).

To create the pixel-level labeling for new images at inference time, we perform the following. The image-level descriptors of the fitting set are clustered using \kmeans to obtain the clusters' centers. Afterward, we apply the feature extractor $F$ on the new images and apply the additional layers $H$ trained using CL. These improved features are then compared with the image-level cluster centers and assigned the label corresponding to the closest point. We apply the analogous steps to compare with the weighted average method of Sohn \etal~\cite{sohn2023anomaly}. 

\section{Discussion on the application domain}
\label{sec:domain}

\begin{figure}[ht]
  \setlength{\tabcolsep}{1pt}%
  \renewcommand{\arraystretch}{0.7}%
  \setlength{\imw}{0.22\linewidth}%
  \newcommand{\vlabel}[1]{%
    \raisebox{0.5\imw}{\rotatebox{90}{\clap{\footnotesize\textbf{#1}}}}}%
  \centering%
  \begin{tabular}{@{}c@{\hspace{1mm}}c@{\hspace{1mm}}c@{\hspace{1mm}}c@{\hspace{1mm}}c@{\hspace{1mm}}c@{}}

    \vlabel{Input image} &
    \frame{\includegraphics[width=\imw]{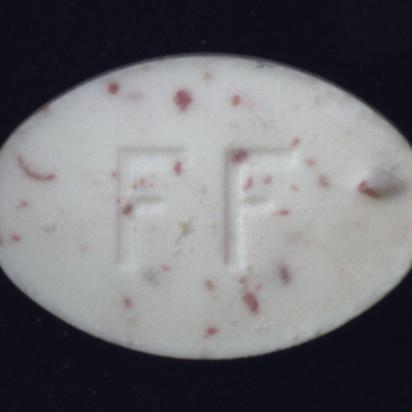}} &
    \frame{\includegraphics[width=\imw]{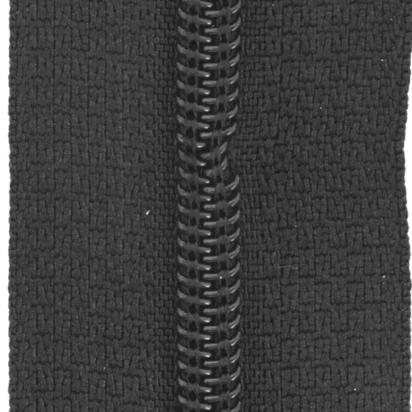}} &
    \frame{\includegraphics[width=\imw]{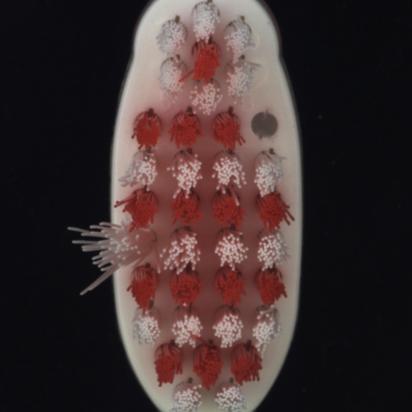}} &
    \frame{\includegraphics[width=\imw]{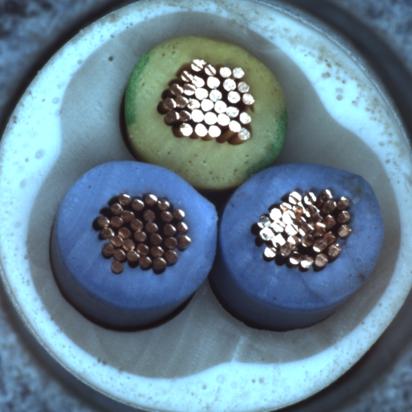}} \\

    \vlabel{GT Mask} &
    \frame{\includegraphics[width=\imw]{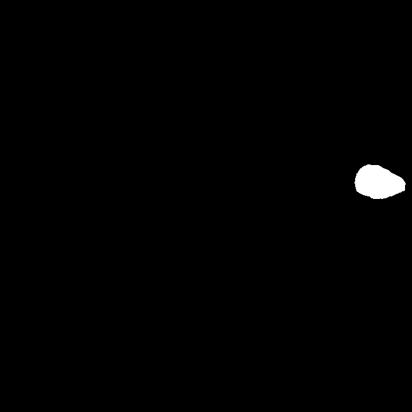}} &
    \frame{\includegraphics[width=\imw]{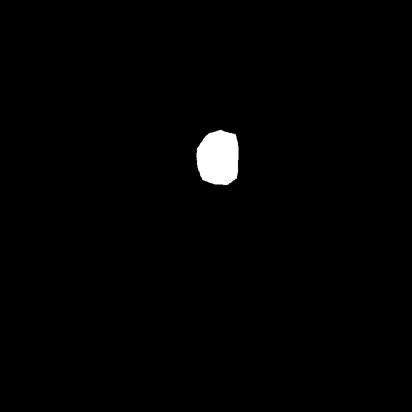}} &
    \frame{\includegraphics[width=\imw]{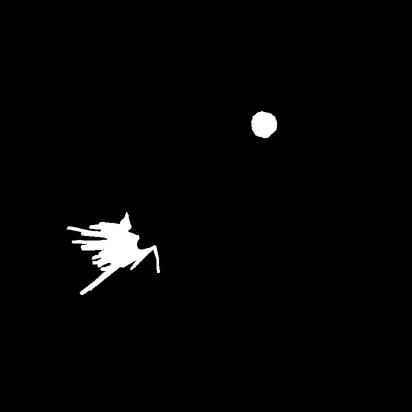}} &
    \frame{\includegraphics[width=\imw]{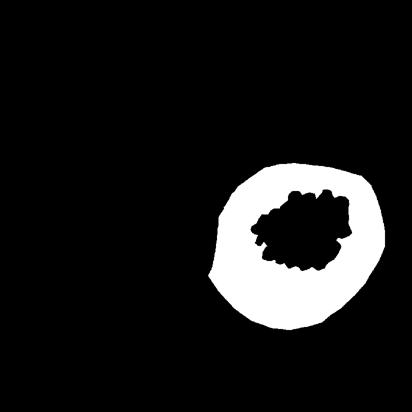}} \\

    \vlabel{Ours} &
    \frame{\includegraphics[width=\imw]{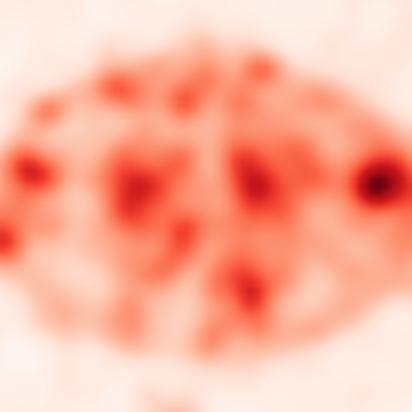}} &
    \frame{\includegraphics[width=\imw]{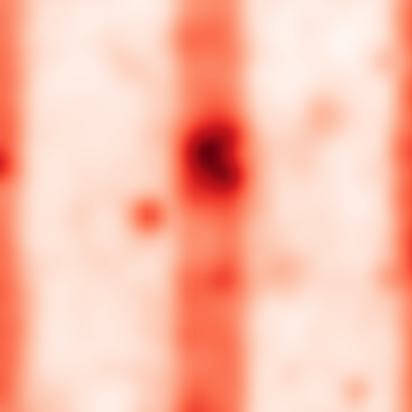}} &
    \frame{\includegraphics[width=\imw]{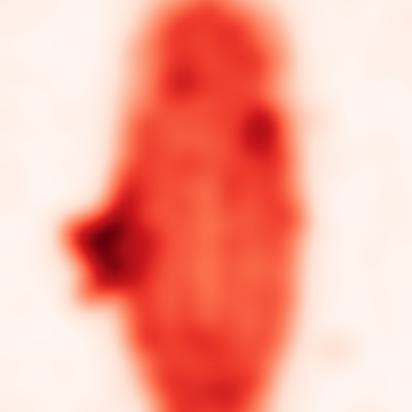}} &
    \frame{\includegraphics[width=\imw]{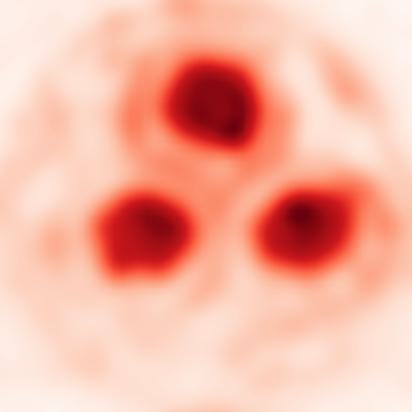}} \\

    \vlabel{FCA} &
    \frame{\includegraphics[width=\imw]{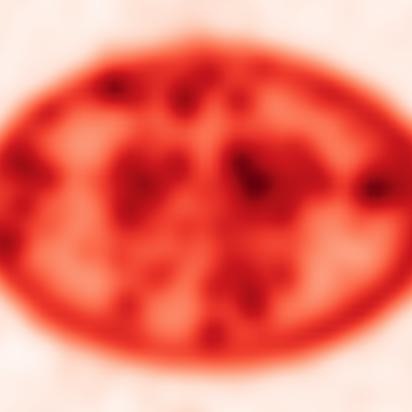}} &
    \frame{\includegraphics[width=\imw]{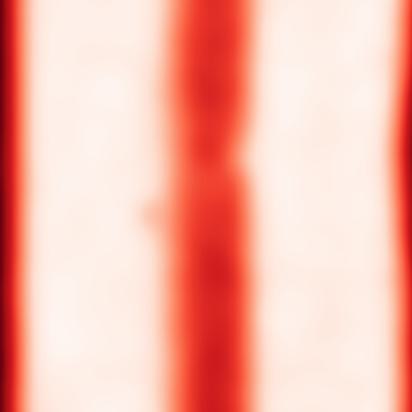}} &
    \frame{\includegraphics[width=\imw]{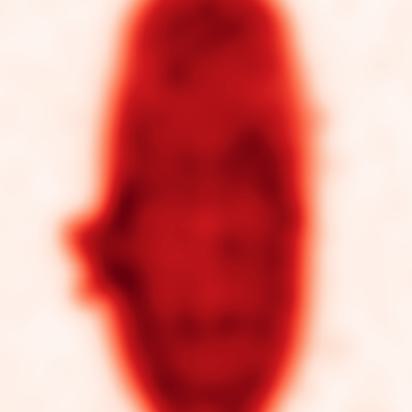}} &
    \frame{\includegraphics[width=\imw]{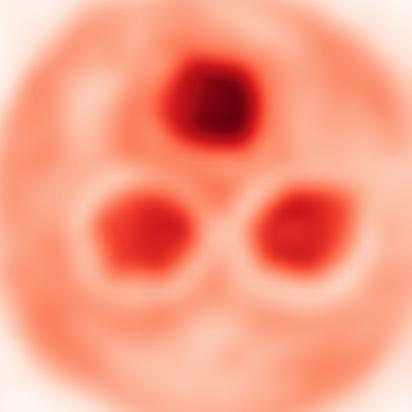}} \\

  \end{tabular}
  \caption{\label{fig:suppl_objects}%
    Qualitative comparison on MVTec AD object classes. The first three examples present structural defects that are reasonably well identified despite the non-homogeneity of the images. In the last column, there is a logical anomaly (wrong cable color) which our method fails to detect.
  }
\end{figure}

As reflected in the MVTec~AD dataset~\cite{bergmann2021mvtec}, there are two different classes of images on which anomaly detection is often applied: textures and generic objects. 
Depending on the class, anomalies tend to manifest themselves in very different ways;
textures contain more subtle, structural anomalies, whereas objects have more semantic or logical anomalies.
Structural anomalies, as found in textures, are better suited for zero-shot~\cite{aota_zero-shot_2023, ardelean2024high} and blind anomaly detection~\cite{zhang2023s}.
Our blind anomaly localization algorithm is designed to work on textures by employing a zero-shot anomaly localization approach (FCA) as a subroutine. Our VAE-based improvement significantly enhances the zero-shot results, especially on more complex textures; however, we still inherit the stationarity assumption of the underlying zero-shot method.
For completeness, we show results on a few objects from MVTec in \Cref{fig:suppl_objects}.
Our blind anomaly localization decidedly improves upon the base zero-shot model.
Nonetheless, the background acts as a distractor and diminishes the capability of the anomaly localization; our approach is also not suited to detect logical anomalies such as the cable swap (last column in \Cref{fig:suppl_objects}).

Importantly, our contrastive learning contribution is complementary to the method used for blind anomaly localization.
If we assume the solution for generating the anomaly maps $A_i$ is given, our feature fine-tuning method can be used to perform anomaly clustering for arbitrary types of images. 

\section{Details on the evaluation of baselines}
In this section, we describe in detail how we obtained the results for the prior work which we compare against in the main paper experiments.

For the main comparison of clustering performance in \Cref{tab:main_results}, we evaluate against seven methods. 
The results for SelFormaly~\cite{lee2023selformaly} are taken from the original paper as no code is shared by the authors. We note that we outperform SelFormaly despite their having a supervisory advantage.
The results for SCAN~\cite{van2020scan} and the approaches introduced by Sohn \etal~\cite{sohn2023anomaly} for MVTecAD textures and MTD are taken from~\cite{sohn2023anomaly}, whereas the metrics for the Leaves dataset we compute ourselves.
We evaluate SPICE~\cite{niu2022spice} using the public official implementation shared by the authors.
To ensure fairness, use the same WideResnet~\cite{BMVC2016_87} feature extractor base for SCAN~\cite{van2020scan} and SPICE~\cite{niu2022spice} as for our method.
In order to evaluate STEGO, which is an unsupervised semantic segmentation method, we must design a mechanism to assign a single label per image. Firstly, we find the most frequently predicted (pixel-level) label and mark it as the normal class. We then use the ground truth number of normal samples ($K$) to the advantage of STEGO by taking the first $K$ images with most pixels assigned to the normal class; these images are labeled as normal. For the rest of the dataset, we assign the image label as the most frequent non-normal label predicted by STEGO.

To evaluate blind anomaly localization (BAL), we compare against the following baselines. 
DRAEM~\cite{zavrtanik2021draem} and CFA~\cite{lee2022cfa} as \semisup methods to analyze the performance when the training set has anomaly contamination ($\approx 75\%$, depending on the dataset).
For convenience, we use the implementation from anomalib~\cite{akcay2022anomalib}.
For ILTM~\cite{patel2023self} and the method of Zhang \etal~\cite{zhang2023s}, we use the results from the respective publications.
ILTM was evaluated in a slightly different way, as the authors used the test set to make train, validation, and test splits and to control the ratio of anomalous samples.
Nonetheless, as Zhang \etal report the results with 80\% anomaly ratio, we argue that their evaluation methodology is largely comparable to ours and the results are representative.
To compare with the anomaly maps generated by the weighted average method of Sohn \etal~\cite{sohn2023anomaly}, we use our implementation of their paper. 
For this experiment, we run their method at a higher resolution ($512 \times 512$) as opposed to the one suggested in the paper ($256 \times 256$) since we observed that this increases the fidelity of anomaly localization.

\section{Code}
The source code for the introduced method is available at \href{https://github.com/TArdelean/BlindLCA}{github.com/TArdelean/BlindLCA}